\documentclass[preprint,11pt]{elsarticle}

\usepackage{geometry}
\usepackage{multirow}
\usepackage{amssymb}
\usepackage{graphicx}
\usepackage{algorithm}
\usepackage{amssymb}
\usepackage{amsmath}
\usepackage{breqn}
\usepackage{multirow}
\usepackage{array}
\usepackage{listings}
\usepackage{csquotes}
\usepackage{booktabs}
\usepackage{caption}
\usepackage{mdwtab}
\usepackage{lipsum} 

\usepackage{soul, xcolor}
\sethlcolor{green}

\journal{Journal Name}

\begin{document}

\begin{frontmatter}

\title{Deep Learning-based Cattle Activity Classification Using Joint Time-frequency Data Representation}

\author[add1,add2]{Seyedeh Faezeh Hosseini Noorbin \footnote{Corresponding author.
\newline{{\it E-mail address:} f.noorbin@uq.net.au (Faezeh Noorbin).}}}

\author[add1]{Siamak Layeghy}
\author[add2]{Brano Kusy}
\author[add4]{Raja Jurdak}
\author[add3]{Greg Bishop-hurley}
\author[add1]{Marius Portmann}

\address[add1]{ITEE, The University of Queensland, Australia}
\address[add2]{DATA61, Commonwealth Scientific and Industrial Research Organisation (CSIRO), Australia}
\address[add3]{Agriculture Flagship Department, Commonwealth Scientific and Industrial Research Organization (CSIRO),Australia}
\address[add4]{Queensland University of Technology, Brisbane, Australia}

\begin{abstract}
Automated cattle activity classification allows herders to continuously monitor the health and well-being of livestock, resulting in increased quality and quantity of beef and dairy products.
In this paper, a sequential deep neural network is used to develop a behavioural model and to classify cattle behaviour and activities. The key focus of this paper is the exploration of a joint time-frequency domain representation of the sensor data, which is provided as the input to the neural network classifier. 
Our exploration is based on a real-world data set with over 3 million samples, collected from sensors with a tri-axial accelerometer, magnetometer and gyroscope, attached to collar tags of 10 dairy cows and collected over a one month period.
The key results of this paper is that the joint time-frequency data representation, even when used in conjunction with a relatively basic neural network classifier, can outperform the best cattle activity classifiers reported in the literature. 
With a more systematic exploration of neural network classifier architectures and hyper-parameters, there is potential for even further improvements. 
Finally, we demonstrate that the time-frequency domain data representation allows us to efficiently trade-off a large reduction of model size and computational complexity for a very minor reduction in classification accuracy. This shows the potential for our classification approach to run on resource-constrained embedded and IoT devices.

\end{abstract}

\end{frontmatter}

\section{Introduction}\label{intro}
Cattle activity classification is of increasing interest to the beef and dairy industry. 
Beef and dairy producers are competing in the international market for greater market share, and are aiming for greater quality and quantity of the production of beef and dairy, as well as reduced cost ~\cite{RN256, RN247}. 
The quantity and quality of beef and dairy product are directly related to the animals' health and welfare, which in turn can be inferred from animal behaviour. Therefore, cattle activity classification has become an increasingly important tool for farmers~\cite{RN253, RN618}. 
For instance, the duration of ``Ruminating", ``Grazing", and ``Lying" can predict productivity, breeding and calving~\cite{RN316}. 
Similarly, changes in gait and duration of ``Lying" can signal pain or lameness in the animal. 
Pain, fear, lameness, heat stress, and infectious disease indicate poor welfare, which is linked to and can be detected via cattle behaviour ~\cite{RN484, RN618}. 
Continuous cattle activity monitoring and classification cannot be done manually (by human observation) at scale, due to its prohibitive cost. This is particularly the case of large scale cattle stations, such as the Brunette Downs and Anne Creek stations in Australia, with 110,000 cattle and an area of 23,677 {$km^2$}, respectively~\cite{RN814, RN262}.

Automated cattle activity classification can be implemented via sensor-based systems, typically equipped with accelerometer, magnetometer and/or gyroscope. These systems have the potential for numerous financial, environmental and other benefits to herders~\cite{RN272, RN106, RN151, RN110, RN155}, as well as for the well-being of cattle~\cite{RN103, RN4}.

There are several ways to approach the cattle activity classification problem. 
Classical Machine Learning methods, referred to as \emph{shallow learning} approaches, have been used to monitor cattle well-being, observe the amount of nutrition intake, and to allocate resources on farms~\cite{RN272, RN106, RN151, RN110, RN155, RN103, RN4}.
Shallow learning classifiers are relatively easy to train, have low computational cost, and work reasonably well in some cases. However, they require extensive feature engineering and are typically limited in the number and types of activities they can accurately classify~\cite{RN131, RN144, RN157}.
Deep Neural Network (DNN) classifiers have the potential to overcome these shortcomings, are able to handle raw sensor readings as inputs, and can outperform shallow learning methods in terms of classification accuracy~\cite{RN131, RN133, RN311}.

In this paper, we explore the use of joint time-frequency data representation (spectrogram) in conjunction with a basic multi-layer perceptron deep neural network for cattle activity classification. 
The rationale behind this approach is the fact that the frequency content of ``bio-signals", such as accelerometry signals from animals and humans, tend to vary over time~\cite{RN479,RN483,fetmov} and is non-stationary. It has been shown that time-frequency data representation is a good fit for such data ~\cite{RN483,RN265,RN162}. This inspires our idea of exploring this type of data representation in the context of cattle activity classification, which is explored in this paper.

As a baseline, we first explore the performance of a basic sequential deep neural network classifier with a time-domain data representation. \footnote{The simple DNN model has also been used for classification by Wang et al.~{\cite{networkref}}} For this, we explore different system parameters, such as data window size,
and degree of window overlap, and we find the respective optimal parameter values. Furthermore, the use and fusion of different sensor modalities (accelerometer, magnetometer and gyroscope) are explored. Our experimental evaluation is based on a manually labelled cattle activity data set with over 3 million samples, collected from 10 dairy cows by CSIRO over a one month period.


We then explore a time-frequency (\textit{spectrogram}) representation of the sensor data in conjunction with the same Neural Network classifier. Our experimental results show a significant classification performance improvement with an $F_1$ Score of $89.3\%$, compared to $73.8\%$ of the time-series data representation approach.
%
We compared our proposed method with the best performing cattle activity classifiers in the literature\footnote{to the best of our knowledge}, including both shallow and deep learning approaches. Our results show that our method outperforms all other reported approaches in terms of $F_1$ Score. 
Our results are achieved for an imbalanced data set, which is a more challenging case than balanced data sets, as used by some of the related works~\cite{RN978}.

It is interesting to note that these results are achieved with a relatively basic multi-layer perceptron DNN classifier, in contrast to some of the considered related works. The key focus of this paper was the exploration of the joint time-frequency data representation, and for that purpose a simple DNN classifier was used. 
The excellent classification result achieved with this data representation, even with a relatively simple classifier, shows its potential. It can be expected that its use in conjunction with more advanced DNN architectures, and a more systematic exploration of the hyper-parameter space, an even higher classification performance can be achieved. This represents a promising avenue for future works.


The paper further considers the computational cost of the time-frequency domain approach, and explores how the change in spectrogram resolution impacts on the classification accuracy. 
Our results show, quite surprisingly, that a significant reduction in the spectrogram resolution by $90\%$, corresponding to a reduction of the model size and computational complexity by $99.9\%$, reduces the classification accuracy by only $1.6\%$. 
Our results demonstrate that a highly compressed spectrogram data representation for a deep learning based cattle activity classifier can achieve very high computational efficiency, while maintaining high classification accuracy.  
This demonstrates the potential for this type of classifier to be run at the edge on very resource-constrained sensor devices, such as the CERES tag~\cite{RN801,RN808}. A more detailed exploration of this promising idea is the focus of our ongoing and future work.

The rest of the paper is organised as follows. 
A brief overview of key related works in cattle activity classification is presented in Section~\ref{CARmethods}. 
Our activity classification architecture and basic approach is presented in Section~\ref{ActivityClassificationArchitecture}, and our data set is discussed in Section~\ref{dataset}. The baseline time-domain-based classification is discussed in Section~\ref{TimeDomainRep}, and Section~\ref{Time-frequency based classification} discusses the corresponding spectrogram-based approach. A detailed performance comparison of the two approaches is provided in Section~\ref{Comptfdtim}, and a comparison with key related approaches is provided in Section~\ref{comparison}.
The trade-off of computational cost versus classification accuracy, via changing the spectrogram resolution, is explored in Section~\ref{compcost}, and conclusions follow in Section~\ref{Conclusion}.

\section{Related Work}\label{CARmethods}
Inferring cattle activity from sensor data using machine learning represents an important and challenging research problem. Our discussion of the key related works in this area is organised according to the two basic types of machine learning approaches that are employed, shallow learning (Section~\ref{CARpSL}) and deep learning (Section~\ref{CARpDL})~{\cite{RN311}}.
Table~\ref{tab:rltdwrk} provides an overview of the key related works in each category, showing the corresponding sensor modalities, machine learning type, classification algorithm, data representation domain, set of activity classes, and classification performance metric ($F_1$ Score).


    %
\vspace{0.15cm}
\begin{table}[t!]

\centering
\caption{Overview of cattle activity classification related works}
\resizebox{\columnwidth}{8.3cm}{
    \begin{tabular}{m{4.355em}llcllc}
    \toprule
    \multicolumn{1}{l}{\textbf{Authors}}  & \multicolumn{1}{m{5.355em}}
    {\textbf{Sensor \newline{} modality}} &  \multicolumn{1}{m{8.285em}}{\textbf{Data \newline{} Representation \newline{} Domain}} & \multicolumn{1}{m{5.285em}}{\textbf{Number of \newline{} Selected \newline{} Features}} & \multicolumn{1}{m{7.355em}}{\textbf{\%Distribution \newline{}in Classes*}} & \multicolumn{1}{m{3.355em}}{\textbf{Classification \newline{}Method}} & \multicolumn{1}{m{7.288em}}{\textbf{Performance\newline{}(\%$F_1$ Score)}} \\
    \midrule
    \multicolumn{1}{l}{\textbf{Shallow Learning}} \\
    \midrule
    \multicolumn{1}{l}{Smith et al.~\cite{RN4}}  & \multicolumn{1}{m{7.355em}}{{\multirow{1}[1]{1.5em}{Accelerometer, \newline{}Orientation}}} & \multicolumn{1}{m{8.9em}}{Time, \newline{}Frequency, \newline{}Information Theory} & 84 & \multicolumn{1}{m{5 em}}{G: 55\newline{}W: 6\newline{}Ru: 13\newline{}Re: 17\newline{}O: 9} &
    \multicolumn{1}{m{8.93em}}{Classification based \newline{}One-vs-all with \newline{}binary classifiers}   & 92.0 \\
    \midrule
    \multicolumn{1}{l}{Rahman et al.~\cite{RN262}}  & {\multirow{1}[1]{7.5em}{Accelerometer}} & \multicolumn{1}{m{8.855em}}{Time, \newline{}Frequency} & 14  & 
    \multicolumn{1}{m{5 em}}{G: 55\newline{}Ru: 16\newline{}S: 29} &
    \multicolumn{1}{m{8.93em}}{Classification based \newline{}One-vs-all with \newline{}binary classifiers} & 91.1 \\
    \midrule
    \multicolumn{1}{l}{Andriamandroso et al.~\cite{RN313}}  & \multicolumn{1}{m{7.355em}}{{\multirow{1}[1]{1.5em}{Accelerometer,\newline{}Gyroscope}}} & Time & 4  & \multicolumn{1}{m{5 em}}{G: 57\newline{}Ru: 4\newline{}O: 39} & 
    \multicolumn{1}{m{7.93em}}{Decision tree} & 80.2 \\
    \midrule
    \multicolumn{1}{l}{Diosdado et al.~\cite{RN315}}   & {\multirow{1}[1]{7.5em}{Accelerometer}} & Time & 2  &
    \multicolumn{1}{m{5 em}}{F: 37\newline{}S: 14\newline{}L: 49} &
    \multicolumn{1}{m{7.93em}}{Decision tree}   & 83.4 \\
    \midrule
    \multicolumn{1}{l}{Arcidiacono et al.~\cite{RN147}} & {\multirow{1}[1]{7.5em}{Accelerometer}} & Time & 1  & \multicolumn{1}{m{7.355em}}{G: 50\newline{}Ru: 50} & \multicolumn{1}{m{7.93em}}{Decision tree} & 94.3 \\
    \midrule
    \multicolumn{1}{l}{Dutta et al.~\cite{RN103, RN106}} &  \multicolumn{1}{m{7.55em}}{{\multirow{3}[1]{7.5em}{Accelerometer,\ Magnetometer,\ Orientation}}} & Time & 72 & \multicolumn{1}{m{7.355em}}{G: 20\newline{}W: 20\newline{}Ru: 20\newline{}Re: 20\newline{}O: 20}& \multicolumn{1}{m{7.93em}}{Ensemble\newline{} classification}   & 90.0 \\
    \midrule
    \multicolumn{1}{l}{Sturm et al.~\cite{RN218}} & {\multirow{1}[1]{7.5em}{Accelerometer}} & Time  & 15 &
    \multicolumn{1}{m{5 em}}{L: 57.8\newline{}S: 42\newline{}W : 0.2} &
    \multicolumn{1}{m{7.93em}}{Chaos-based \newline{}classification} & 87.9 \\
    \midrule
    \multicolumn{1}{l}{Martiskainen et al.~\cite{RN821}}  & {\multirow{1}[1]{7.5em}{Accelerometer}} & Time & 28 
    & \multicolumn{1}{m{5 em}}{S: 22\newline{}L: 19\newline{}Ru: 20\newline{}F: 17\newline{}W: 9\newline{}LW: 9\newline{}L: 1\newline{}S: 2} & \multicolumn{1}{m{7.93em}}{SVM}    & 77.7 \\
    \midrule
    \midrule
    \multicolumn{1}{l}{\textbf{Deep Learning}} \\
    \midrule
    \multicolumn{1}{l}{Kasfi et al.~\cite{RN221}}  & {\multirow{1}[1]{7.5em}{Accelerometer}} & Time & \textbf{\_\_} & 
    \multicolumn{1}{m{5 em}}{G: $\sim$ 60\newline{}O: $\sim$40} & \multicolumn{1}{m{7.93em}}{ConvNet \newline{}with Softmax} & 84.0 \\
    \midrule
    \multicolumn{1}{l}{Rahman et al.~\cite{RN263}}  & {\multirow{1}[1]{7.5em}{Accelerometer}} & Time  & \textbf{\_\_} & 
    \multicolumn{1}{m{5 em}}{G:11.11 \newline{}Se:11.11 \newline{}W:11.11 \newline{}C:11.11\newline{}RuL:11.11\newline{}RuS:11.11\newline{}ReL:11.11\newline{}ReS:11.11\newline{}O:11.11 } & \multicolumn{1}{m{7.93em}}{Auto-encoder \newline{}with SVM} & 75.1 \\
    \midrule
    \multicolumn{1}{l}{Peng et al.~\cite{RN131}}  & \multicolumn{1}{m{7.355em}}{{\multirow{1}[1]{7.5em}{Accelerometer, \newline{}Magnetometer,\newline{}Gyroscope}}} & Time &\textbf{\_\_} & 
    \multicolumn{1}{m{5 em}}{F:12.5\newline{}M: 12.5\newline{}LS: 12.5\newline{}RuS: 12.5\newline{}RuL: 12.5\newline{}L: 12.5\newline{} SL: 12.5 \newline{}H:12.5 }&
 \multicolumn{1}{m{7.93em}}{RNN-LSTM \newline{}with Softmax}  & 88.9 \\
    \midrule
    \multicolumn{7}{p{61.495em}}{* C: Chewing, F: Feeding, G:Grazing, H: Head butt, L: Lying, LS: Licking Salt,  LW: Lame Walking, \newline{}M: Moving, O: Other,  Re: Resting, Ru: Ruminating, S: Standing, Se: Searching, SL: Social Licking, W: Walking,} \\
    \bottomrule
    \end{tabular}%
  \label{tab:rltdwrk}
  }
  
\end{table}%

\subsection{Shallow Learning Methods for Cattle Activity Classification}\label{CARpSL}
Currently, the majority of studies dealing with cattle activity classification use shallow learning, i.e. conventional machine learning methods, to build behavioural models of cattle activity classes.  
%
%
%
Smith et al. used a one-vs-all classification method, where a separate binary classifier is used for each activity class \cite{RN4}. 
The authors propose an embedded feature selection method for each of the binary classifiers that is tailored to the particular activity, with a specific optimal window size. For example, the authors show that a Random Forest Ensemble classifier with a 30 sec time window is optimal for ``Grazing", while a SVM-based classifier with a 20 sec time window is more efficient for ``Walking".
%
%
%
%
The authors use a tri-axial accelerometer and extract 84 features in the time and frequency domains, in addition to information theory-based features such as Entropy.
The authors report an average $F_1$ Score of $92.0\%$ for five activity classes (``Grazing", ``Walking", ``Ruminating", ``Resting" and ``Other"). 


In~\cite{RN262}, Rahman et al. apply a similar classification method on tri-axial accelerometer data, using 14 time and frequency domain features.
The authors use majority voting to integrate the different binary classifier outputs.
The results are reported for the three most dominant activity classes with an average $F_1$ Score of $84.4\%$ from sensor data collected from collar tags, and $91.1\%$ for data collected from halter tags. Both data sets were highly imbalanced. 
The above two methods only deal with a relatively small number of activity classes. 
Furthermore, running all the class-optimal classifiers in parallel results in a relatively high computational and hence energy cost. 


Andriamandroso et al.~\cite{RN313} proposed a decision tree-based activity classification algorithm using 4 selected features derived from accelerometer and gyroscope signals, created mainly via head and jaw movements of the cattle. 
These movement signals are collected using the inertial measurement unit (IMU) of an iPhone that is attached to the animals, along with a backup battery. 
The average $F_1$ Score in this study is reported as $80.2\%$, for the activity classes of ``Grazing", ``Ruminating" and ``Other" (including all other activity classes), using an imbalanced dataset.

Diosdado et al.~\cite{RN315} proposed a similar approach based on two selected features from acceleration signals collected from collar tags. The classification performance is reported as $83.4\%$ (average $F_1$ Score), but only three activity classes were considered (``Lying", ``Standing" and ``Feeding"), based on an imbalanced dataset.


In ~\cite{RN147}, Arcidiacono et al. employ a simple decision tree algorithm for a binary activity classification problem, with ``Feeding" and ``Standing" as the only two considered classes. In this approach, the median of the X-axis acceleration values over a 5 second window is used as the key feature and decision threshold. If fewer than 10 data samples in the 5 second time window are below the threshold, the window is classified as activity class ``Feeding", and ``Standing" otherwise. 
The authors report an $F_1$ Score of $94.3\%$ for this binary classification problem. The main issue with this simple threshold-based approach is its limited ability to scale to a larger number of activity classes.


Another shallow learning approach, proposed by Dutta et al. in ~\cite{RN106, RN103}, is an ensemble classifier which combines the outputs of several individual classifiers. 
The authors considered 72 time-series features collected from an IMU sensor. 
The paper reports an $F_1$ Score of $90\%$ for five activity classes (``Grazing", ``Walking", ``Ruminating", ``Resting" and ``Other").

In ~\cite{RN218}, Sturm et al. used a nonlinear dynamical system model and chaos theory based approach to model the problem cattle activity classification. 
%
%
Seven chaos-theory related features, such as the Maximal Lyapunov exponent and the length of a phase trajectory, were extracted from tri-axial acceleration sensor data.
The $F_1$ Score of the proposed method for up to three activity classes is $87.9\%$, using an imbalanced dataset. However the classification performance decreases significantly for the case where six activity classes are considered. 

In~\cite{RN821} Martiskainen et al. propose a SVM-based classifier for eight cattle activity classes. The authors consider 28 time-domain features from tri-axial acceleration sensors, and use a grid search method to find the optimal SVM hyper-parameters. The $F_1$ Score of this approach for eight activity classes is $77.7\%$, with an imbalanced dataset.

Overall, we can say that the discussed shallow learning approaches for cattle activity classification can provide very good classification performance for a small number of activity classes, but the performance decreases significantly if a larger number of activity classes (more than 5) is considered.  
In the following, we will discuss the key works that have considered deep neural networks to solve the cattle activity classification problem.

\subsection{Deep Learning Methods for Cattle Activity Classification}\label{CARpDL}
Deep neural networks have proven to be very successful for a wide range of classification problems, and so it does not come as a surprise that they have been explored for cattle activity classification. A range of deep learning methods, such as Convolutional Neural Networks (ConvNet), Auto-encoders, and Long Short Term Memory Recurrent Neural Networks (RNN-LSTM), have been explored for activity classification~\cite{RN477}. Below, we provide a brief summary of the key works in the context of cattle activity classification.


Kasfi et al.~\cite{RN221} employ a ConvNet with a Softmax classification layer, and consider accelaration sensor data in time-series representation. 
The method achieves an $F_1$ Score of $84\%$ for binary classification of ``Grazing"  versus other activities, with a reasonably balanced dataset.

In~\cite{RN263}, Rahman et al. propose the use of a stacked Auto-encoder for feature representation of time-domain acceleration sensor data, used in conjuction with an SVM-based activity classifier.
The paper reports an $F_1$ Score of $75\%$ for 9 activity classes. 

Peng et. al.~\cite{RN131} employ an RNN-LSTM algorithm to represent features based on time-series sensor data from a 9-axis Inertial Measurement Unit (IMU). 
A final Softmax layer provides the classification functionality. The paper considers eight activity classes and the proposed classifier achieves an average $F_1$ Score of $89\%$, which to the best of our knowledge, is the highest reported performance for such a large number of activity classes. However, this result was achieved on a balanced data set, with a roughly equal representation of the different activities. Realistic data sets tend to be imbalanced, which makes the classification problem harder, and typically results in lower classification performance. As a final caveat, the RNN-LSTM classifier is relatively resource intensive and requires a high memory bandwidth~\cite{Chang2015}, which makes it difficult to be deployed on resource-constrained edge devices.

Our research presented in this paper is motivated by our review of the relevant literature, and our aim was to develop a new cattle activity classification method that combines high classification performance for a large number of activity classes, with low and scalable computational complexity, and with the ability to work with realistic (imbalanced) data sets.

\section{Dataset}\label{dataset}
The dataset used in this paper has been collected by DATA61 (CSIRO) in Armidale, Australia, from the 31st of July to 4th of September 2018. The dataset consists of $\sim3,500,000$ samples of tri-axial IMU sensor readings from an accelerometer, magnetometer and gyroscope, attached to collar tags of ten individual dairy cows. 
Given a sampling rate of 50 Hz, the dataset corresponds to a duration of $69,150$ seconds, or just over 19 hours.


The dataset has been labelled with the corresponding cattle activity class by human recorders. The considered set of activity classes are: ``Grazing", ``Walking", ``Ruminating standing", ``Ruminating lying", ``Standing", ``Lying", ``Drinking", ``Grooming", and ``Other". 
%
Table{~\ref{tab:d1} shows} the respective aggregate activity duration in seconds as well as the distribution in percent.
We can see that the dataset is highly imbalanced, with over 
53\% of data samples labelled with the ``Grazing" activity label.

\vspace{15pt}
\begin{table}[t]
\small
\centering
\caption{Dataset breakdown by activity class.}
\small
\begin{tabular}{clcc}
    \toprule
    \multicolumn{1}{p{3.5em}}{Label} & \multicolumn{1}{l}{Activity class}  & \multicolumn{1}{p{6.85em}}{Duration (s)} & \multicolumn{1}{p{7.85em}}{Distribution (\%)} \\
    \midrule
    1     & Grazing  & 36701 & 53.07 \\
    2     & Walking  & 514   & 0.74 \\
    3     & Ruminating standing  & 3392  & 4.91 \\
    4     & Ruminating lying  & 9851  & 14.25 \\
    5     & Standing  & 11471 & 16.59 \\
    6     & Lying  & 5426  & 7.85 \\
    7     & Drinking & 611   & 0.88 \\
    8     & Grooming & 530   & 0.77 \\
    9     & Other  & 654   & 0.95 \\
    \bottomrule
    \end{tabular}%
  \label{tab:d1}%
\end{table}%


\section{Deep Learning based Activity Classification Architecture}\label{ActivityClassificationArchitecture}

\begin{figure}[!ht]
\centerline{\includegraphics[width=\textwidth]{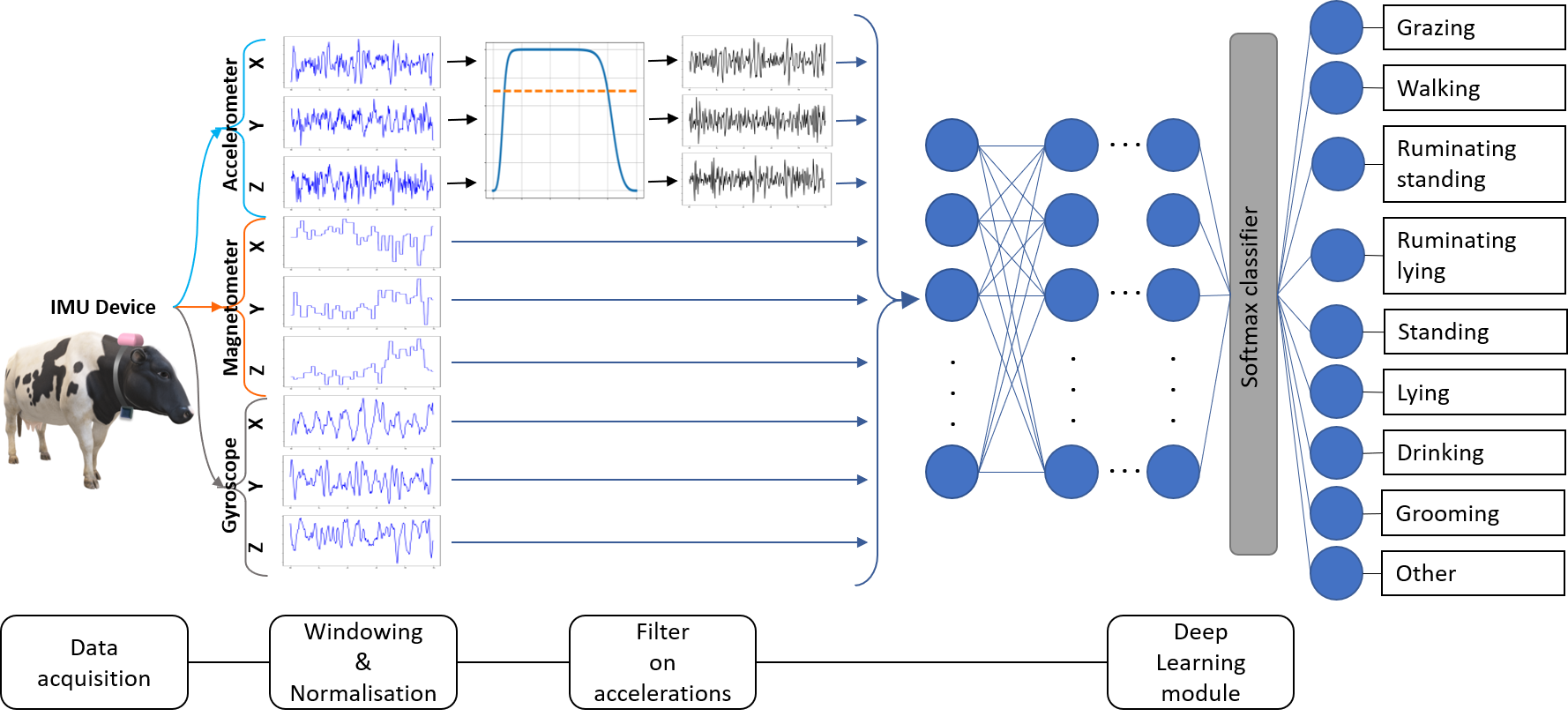}}
\caption{Cattle Activity Classification Architecture.}
\label{fig:Architecture}
\end{figure}

Figure~\ref{fig:Architecture} shows an overview of our basic Deep Learning based activity classification architecture and approach. The basic idea and very preliminary results have been published in{~\cite{myPhdForum}}. The process starts on the left, with data acquisition from the tri-axial IMU sensor that is attached to the animal. This provides raw data for the three dimensions from each sensor (gyroscope, magnetometer and accelerometer), resulting in nine time-series, as indicated in the figure. This is then followed by windowing and data normalisation, as well as filtering applied to the accelerometer data, in order to mitigate the gravity effect.  The data is then fed into our Deep Learning based classifier, which outputs one of the nine considered activity classes, including ``Other", as mentioned in  Section~\ref{dataset}. These individual processing steps are discussed in more detail in the remainder of this section.

\subsection{Windowing and Normalisation}\label{Pre-processing}
\begin{figure}[!ht]
\centerline{\includegraphics[width=9.3cm]{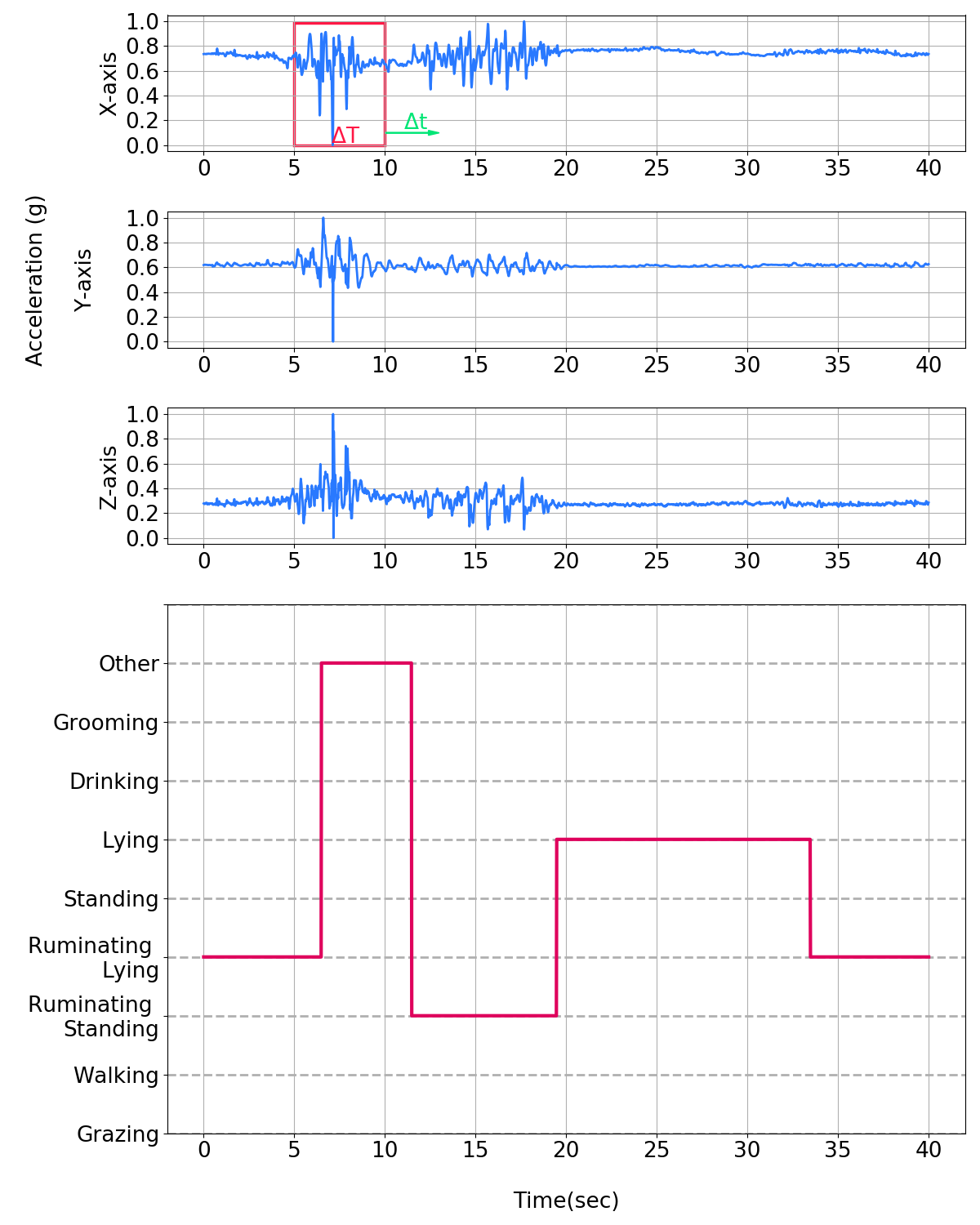}}
\vspace*{8pt}
\caption{Acceleration time-series data on x, y, and z axes, with corresponding activity labels (below).}
\label{fig:TSsignalnLabeltrace}
\end{figure}

Since cattle activities cannot be determined based on instantaneous sensor readings, we need to consider the appropriate data window size, which the classifier considers in each classification step.   
To illustrate this, Figure~\ref{fig:TSsignalnLabeltrace} shows the collected tri-axial accelerometer time-series data for a period of 40 seconds. The graph at the bottom shows the corresponding activity classes, based on the manually labelled data set. We can see that the shortest activity duration in this example is around 5 seconds, while others last significantly longer. 
In order to implement reliable activity classification, the choice of window size is important.
We explored the following values for the window size $\Delta T$, i.e. $\Delta T\in \{\textrm{5s},  \textrm{10s},  \textrm{15s}\}$.

We also explored a range of values for the  window overlap percentage $P$. 
For example, an overlap of $P=40\%$ for a window size of $\Delta T = 5s$ corresponds to a stride of $\Delta t = 2s$ for each classification cycle.
We considered the following values ${P} \in\{0\%, 40\%, 80\%\}$ for each considered window size.
\footnote{In Figure~\ref{fig:TSsignalnLabeltrace} the window size $\Delta T$  is shown in red, and the stride $\Delta t$ is indicted in green.}
Finally, the sensor readings are normalised to a range between 0 and 1 using the Min-Max method.
%
%





\subsection{Filtering}
\begin{figure}[!ht]
\centerline{\includegraphics[width=7.5 cm]{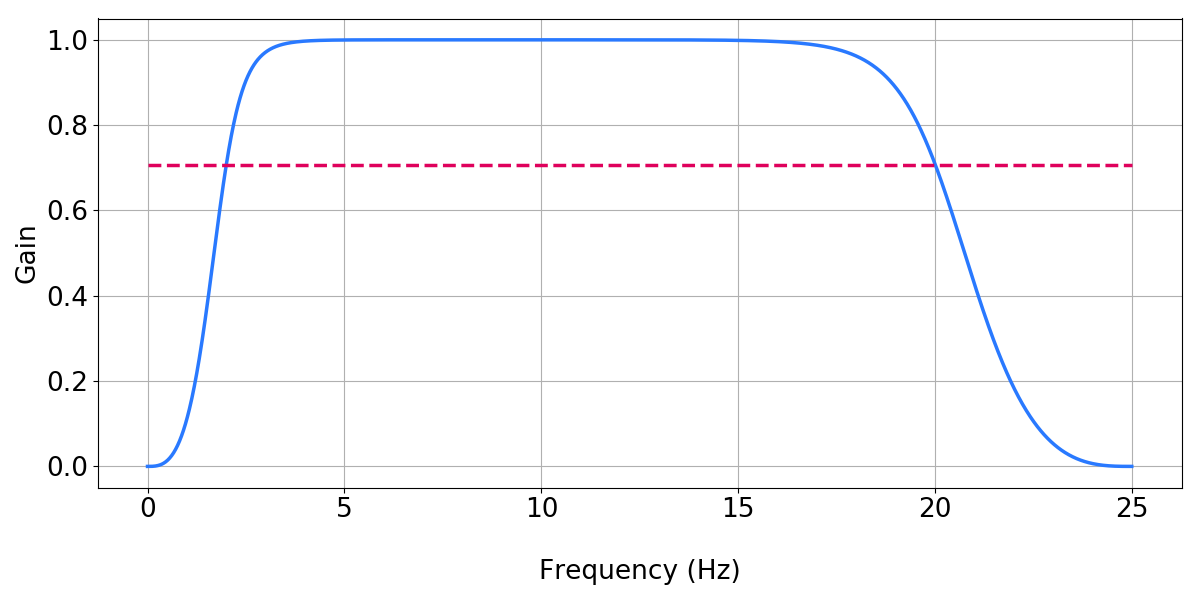}}
\vspace*{4pt}
\caption{Frequency response of filter.}
\label{fig:FrqRspnsFilt}
\end{figure}

The tri-axial accelerometer readings include not only the acceleration generated by different cattle activities, but also the impact of gravity as well as thermal noise~\cite{RN826}.
In order to address this problem, we apply a bandpass filter to the accelerometer data, with the aim to eliminate the gravity induced DC component, as well high frequency noise ~\cite{RN272}.
Based on a set of experiments, we designed a third-order Butter-worth bandpass filter with a lower cut-off frequency of $2 Hz$ and higher cut-off frequency of $20 Hz$. 
The frequency response of the filter is shown in Figure~\ref{fig:FrqRspnsFilt}.



\subsection{Deep Neural Network Module}\label{DNN}
\begin{figure}[!ht]
\centerline{\includegraphics[width=9.5cm]{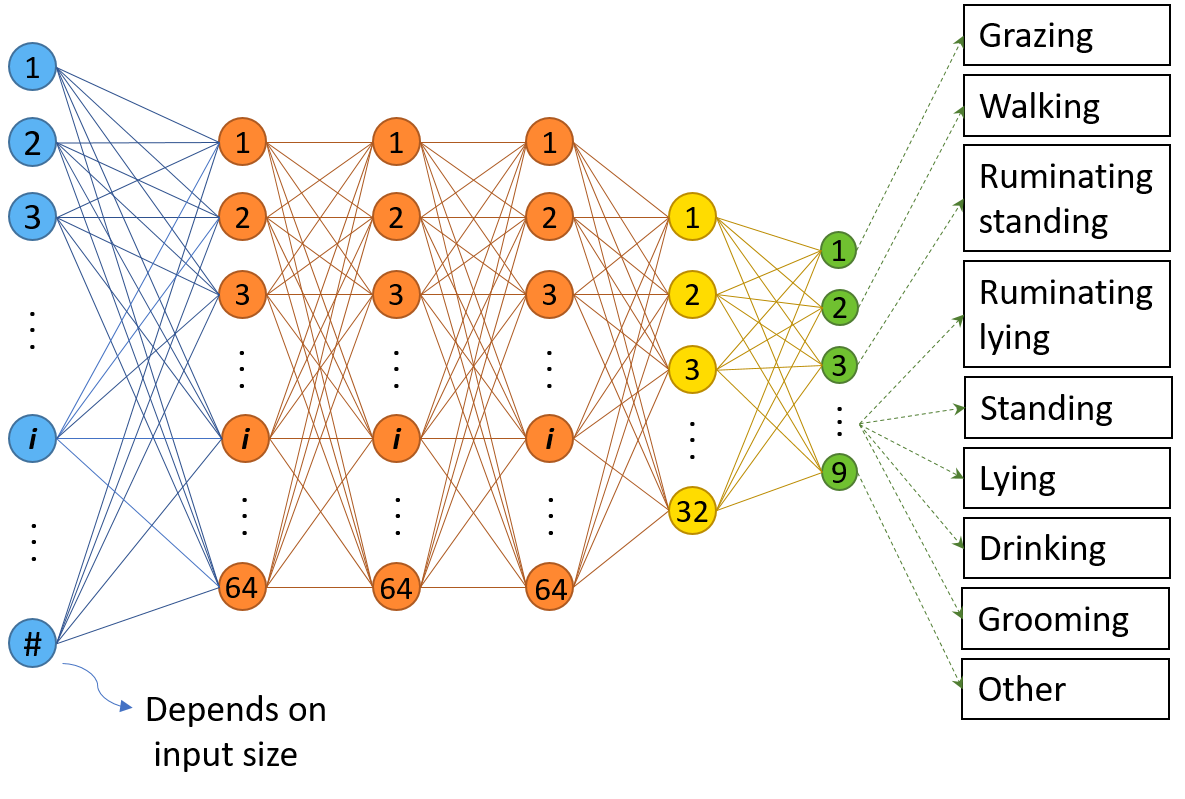}}
\vspace*{8pt}
\caption{A schematic illustration of DNN module.}
\label{fig:netmodel}
\end{figure}

In the basic version of our cattle activity classifier, the time-series sensor data is fed into a feed forward Deep Neural Network (DNN)  classifier, after windowing, normalisation and filtering has been applied.
Figure~\ref{fig:netmodel} shows the structure of the DNN module, comprising a stack of fully connected successive layers, consisting of an input layer, followed by 4 hidden layers, and the output layer. 
The number of nodes in the first layer depends on the number of data samples in each window, and hence the window size and number of sensor modalities.
The number of nodes in the hidden layers are 64, 64, 64 and 32 respectively. The output layer has 9 nodes, corresponding to the 9 activity classes. 

The Rectified Linear Unit (ReLU) activation function was used for all layers except the final one, where the Softmax function was chosen. This choice of ReLU and Softmax activation function is a common and proven combination for a wide range of classification problems~\cite{RN464,RN465, RN466}. 
Finally, Sparse Categorical Cross-entropy~\cite{Murphy2012} was chosen as the loss function. 
 


\subsection{Implementation and Training of Time-domain DNN Model}\label{validation time}
The neural network model discussed in the previous section was implemented using Keras (2.2.4-tf) and Tensorflow(2.0.0) ~\cite{keras,tensorflow2015}. For the training of the network, we used the Adam optimiser~{\cite{kingma2014}} with parameters $\beta_1$, $\beta_2$, $\epsilon$, and learning rate set to 0.9, 0.999, $10^{-7}$, and 0.001, respectively. The number of epochs ranged between 2 to 10 to get the best fit of the models, and the batch size was 32.
As a validation method, we used Stratified K-fold Cross Validation (SCV) with $K = 10$~\cite{RN827}.
We compute the overall performance as the weighted average $F_1$ Score over the different classes.

\section{Activity Classification Results using Time-domain Data Representation}\label{TimeDomainRep}

In this section, we discuss the application of our Deep Learning based cattle activity classification approach, based on sensor data in time-domain representation. As discussed in the previous section, the time-series sensor data undergoes windowing, normalisation and filtering, before being presented to the DNN classifier. We first explore different window sizes and degrees of window overlap (or stride), as discussed in Section~\ref{Pre-processing}.
We then present the classification performance results for different individual sensor modalities, as well as the different options of sensor modality fusion.


\subsection{Exploring Window Size and Overlap}\label{restimewindow}
\begin{figure}[!ht]
\begin{center}
\includegraphics[width=13.5cm]{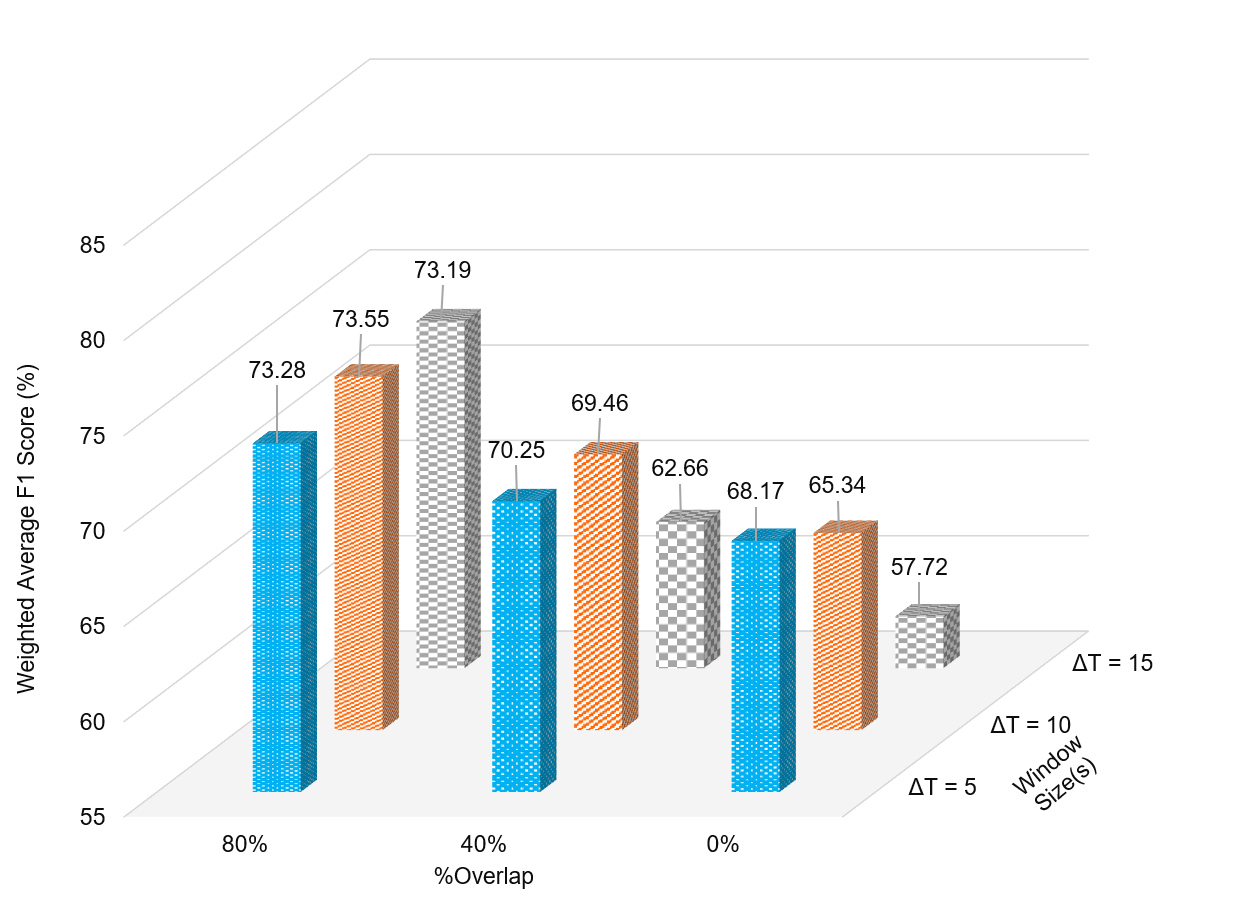}
\caption{\small Classification performance for time-series data representation  with varying window overlap.}
\label{fig:diffWin}
\end{center}
\end{figure}

In our first experiment, we trained our DNN classifier, as discussed in Section~{\ref{validation time}}, using only acceleration time-series data. We considered three values for the window size ($\Delta T$), i.e. 5, 10 and 15 seconds. For each window size, we considered different degrees of window over window overlap, i.e. 0\%, 40\% and 80\%. 
Figure~\ref{fig:diffWin} shows the classification performance in terms of the weighted average $F_1$ Score, for all nine combinations of window size and overlap.
The x axis of the figure represents the different window overlap values, while the  y axis represents the different window size options. The weighted average $F_1$ Score is represented on the z axis.

We observe that the choice of window size and overlap has a significant impact on the classification performance, ranging from a minimum of $51.72\%$ for $\Delta T = 15s$ and overlap of $0\%$, to a maximum $F_1$ Score of $73.55$ for $\Delta T = 10s$ and a window overlap of $80\%$.
In general, a greater window overlap seems to result in a better performance for all window sizes. This can be attributed to information loss and artefacts at the window edges.
For the best overlap value of $80\%$ the choice of window size does not seem to have too much of an impact. 

\subsection{Exploring Sensor Modalities}

In a next set of experiments, we explored the use of different combinations of sensor modalities for our time-domain based DNN classifier. For these experiments, we chose the two best options of window size and overlap/stride, as established in Section~\ref{restimewindow}, i.e. $\Delta T = 10s$ and $\Delta T = 5s$, with a stride of $\Delta t = 2s$ and $\Delta t = 1s$ respectively, which both correspond to a window overlap of 80\%.
%
The sensor modality combinations that we considered are: accelerometer data only (as in Section ~\ref{restimewindow}), accelerometer \& magnetometer, accelerometer \& gyroscope, and accelerometer \& magnetometer \& gyroscope. 
Data from the different modalities were combined by simply concatenating the various tri-axial sensors data values. 

\begin{figure}[!ht]
\centerline{\includegraphics[width=11cm]{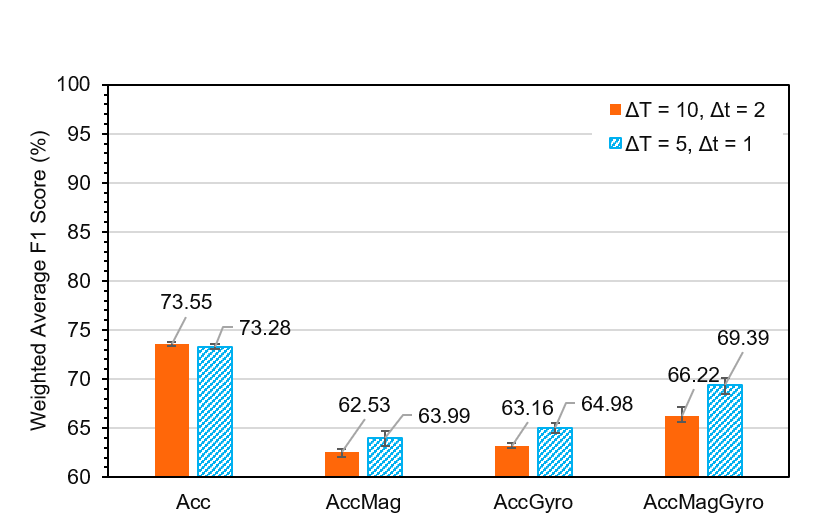}}
\vspace*{8pt}
\caption{Classification performance for time-series data representation for different sensor modality sets.}
\label{fig:TimeFusionSensors}
\end{figure}

Figure~\ref{fig:TimeFusionSensors} shows the result of our experiments. As discussed in Section~\ref{restimewindow}, classification based on accelerometer data only achieves an $F_1$ Score of 73.28\% and 73.55\% respectively. What is surprising is that the addition of Magnetometer data deteriorates the performance, down to 62.53\% and 63.99\% respectively.  The same applies for the case where we combine acceleration with gyroscope data, and the case where we combine all three sensor modalities. 
It seems that the classifier is not able to utilise the additional information from the different sensor modalities, if provided in time-domain representation.

\begin{figure}[!ht]
\centerline{\includegraphics[width=\textwidth/2]{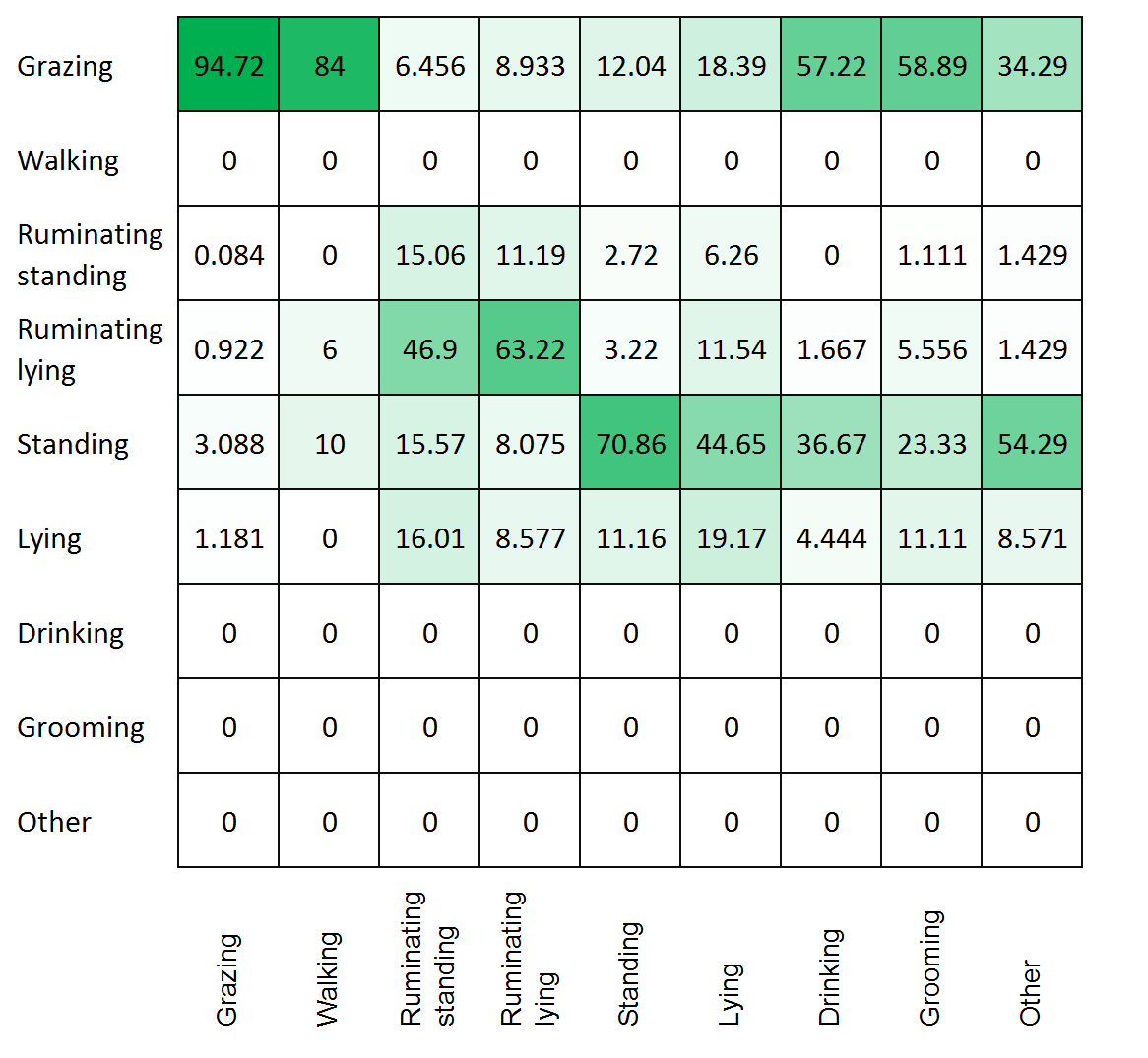}}
\vspace*{8pt}
\caption{Confusion matrix for classification based on time-domain data representation, for 
$\Delta T = 10s$ and $\Delta t = 2s$, based on accelerometer time-series data only,  using the SCV method.}
\label{fig:TimeConfusionMat}
\end{figure}

Figure~\ref{fig:TimeConfusionMat} shows the confusion matrix for the best performance of time-domain-based classification with $\Delta T = 10s$ and $\Delta t = 2s$, using accelerometer time-series data with the Stratified 10-fold cross-validation (SCV) method. 
The values in a column indicate the percentage of the predicted individual class to the total true class. 
We can see that, while  $94.72\%$ of the samples in the ``Grazing" class are correctly classified, a large percentage of samples of other activities are misclassified.
For example, the classification performance for the ``Walking" class is extremely poor, with $0\%$ of the labels predicted correctly, and with a high degree of confusion with ``Grazing".

It is clear from our results that the performance of  a simple time-series data representation in combination with a basic DNN classifier does not achieve a sufficiently high performance for most practical applications. 
This provides the motivation of our exploration of an alternative data representation. 
Our choice of data representation is based on the fact that the frequency content of many ``bio-signals" have been shown to vary over time. This has been shown to be the case for animals~\cite{RN479}, accelerometry signals of human fetal activity~\cite{fetmov}, and human activity signals more generally~\cite{RN483,RN265,RN162}. 
Our decision to explore a joint time-frequency sensor data representation (spectrogram) for cattle activity classification was inspired by these related works.

The following section provides details of our spectrogram-based sensor data representation and the application in the context of DNN cattle activity classification. \footnote{Since our focus in this paper is on exploring sensor data representation, rather than exploring optimal neural network architectures and hyper-parameters,  we use a relatively simple DNN classifier with a basic set of ``default" parameters.}




\section{Activity Classification with joint Time-Frequency Data Representation}\label{Time-frequency based classification}
The cattle activity classification performance using a DNN classifier based on the time-series data representation, as shown in the previous section, is limited and not sufficient for practical purposes. In this section, we explore a time-frequency data representation approach, and demonstrate how it significantly improves classification accuracy using the same DNN classifier. 



\subsection{Time-Frequency Data Representation}


A commonly used alternative representation of raw time-series data is a frequency domain representation, e.g. via the \textit{Fourier} transform.

In Fourier analysis, we use a weighted average of the signal in the time domain, which is mathematically expressed for aperiodic signal $x(t)$ as:
\begin{flalign}
X(f)=\int_{-\infty}^{\infty} x(t)e^{-2j\pi ft} dt
\label{formula:fourier}
\end{flalign}

The Fourier transform, in general, is a complex function of frequency only, i.e. the time variable is integrated out. While the magnitude of the Fourier transform indicates the harmonics/frequencies present in a signal, it does not provide information relating the \textit{time of arrival} of individual frequencies. 
Many human-made and natural signals have been shown to have varied harmonic components over time, including human speech, bird sounds,  bio-signals of human and animals~\cite{RN479} such as accelerometry signals of human fetal activity~\cite{fetmov}, and more closely related to our work, human activity~\cite{RN162,RN483}. 
For a complete harmonic analysis of these \textit{non-stationary} signals, i.e. signals with their frequency/harmonic content varying over time, methods are needed that represent signal (harmonics) in a \textit{joint time-frequency} domain. 
There are different types and classes of time-frequency data representations that, each with their own advantages and disadvantages. 

The method used in our work is based on the \textit{Short-Time Fourier Transform (STFT)}, which is a localised form of the Fourier transform that calculates the Fourier transform over a small sliding window of the signal.
For this, the time-series signal  $x(t)$ is first multiplied by a windowing function  $W(t)$, such as the \textit{Hamming} function,  with $\tau$ representing the time shift of the window. 

\begin{flalign}
x_W(t,\tau)=x(t)W(t-\tau)
\label{formula:STFT1}
\end{flalign}

The STFT is then calculated as the Fourier transform of $x_W(t,\tau)$ with the integral taken over the time shift variable $\tau$. 

\begin{flalign}
X_W(t,f)&=\int_{-\infty}^{\infty} x_W(t,\tau)e^{-2j\pi f\tau} d\tau \\ 
&=\int_{-\infty}^{\infty} x(t)W(t-\tau)e^{-2j\pi f\tau} d\tau
\label{formula:STFT2}
\end{flalign}

The STFT $X_W(t,f)$ is a complex function that represents the signal both in terms of time and frequency. 
The STFT belongs to a more general family of \textit{Time-Frequency Distributions (TFD)}, and hence we will use the term TFD for this type of data representation.   
%
In this paper, we use the square magnitude of STFT, which is typically referred to as \textit{spectrogram}, since it is able to represent the signal energy distribution over the joint time-frequency domain. 

The spectrogram of a signal $x(t)$ is calculated as follows:

\begin{flalign}
S^{(W)}_{x}(t,f)&=|X_W(t,f)|^2 \\
&=|\int_{-\infty}^{\infty} x(\tau)W(\tau-t)e^{-2j\pi f\tau} d\tau|^2
\label{formula:STFT3}
\end{flalign}

\begin{figure}[!t]
\centerline{\includegraphics[width= \textwidth]{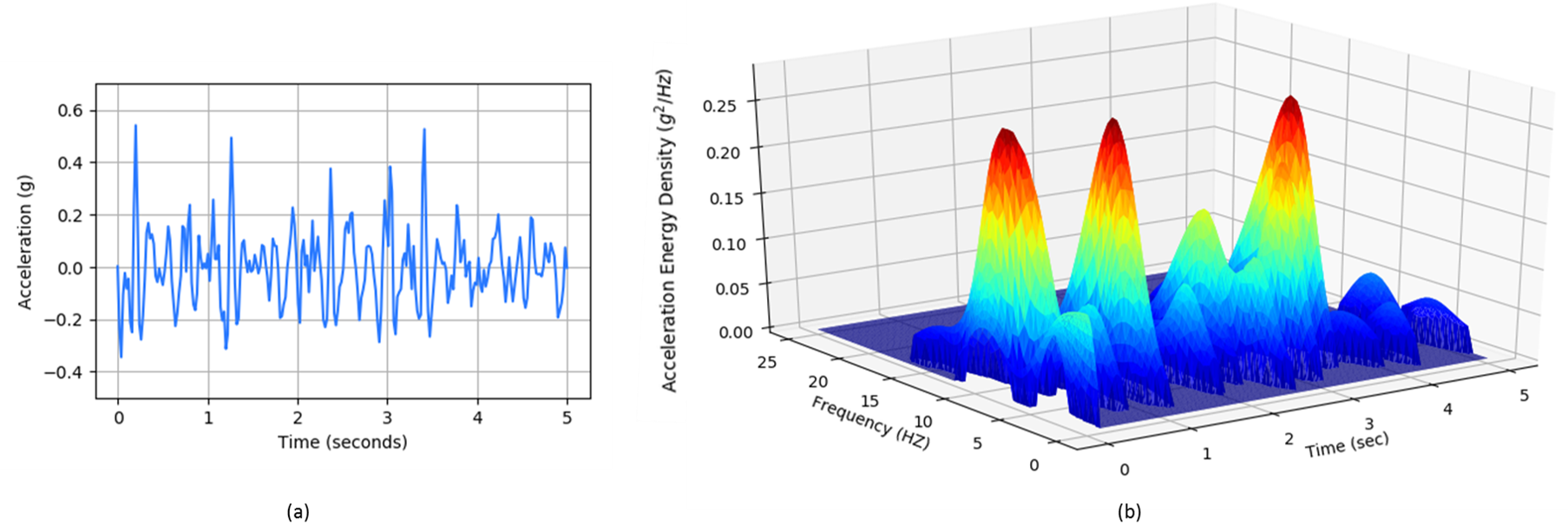}}
\vspace*{4pt}
\caption{Sample window of cattle activity acceleration signal (z axis) in time-domain representation (a), and its corresponding TFD (spectrogram) representation (b).}
\label{fig:TSnTFD3d}
\end{figure}

\noindent Figure \ref{fig:TSnTFD3d} shows an example window of the cattle activity acceleration signal on the left, and its corresponding spectrogram visualised in 3D on the right. The $x$ and $y$ axes indicate the time and frequency respectively, and the $z$ axis indicates the acceleration energy density in $g^2/Hz$ (also indicated by colour).



%

\begin{figure}[!t]
\centerline{\includegraphics[width=10.5 cm]{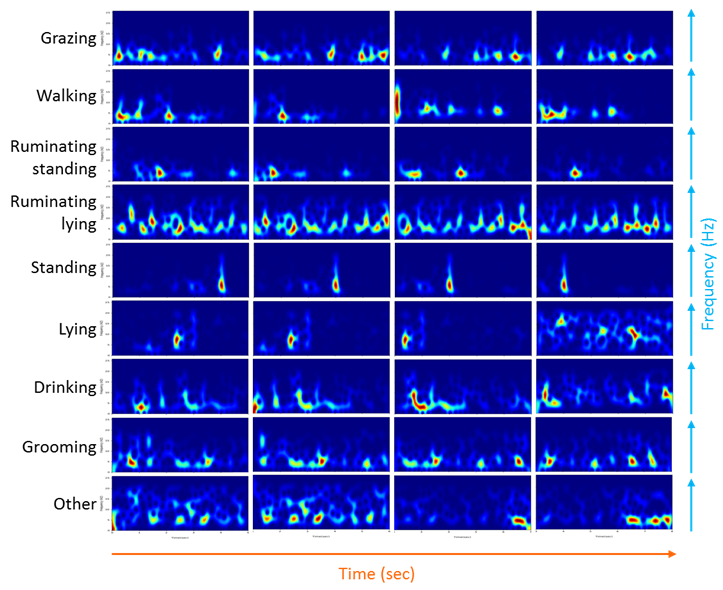}}
\vspace*{4pt}
\caption{Spectrograms representation of eight consecutive seconds of acceleration signal (z axis) for different activity classes.
}
\label{fig:TFDdiffLabel}
\end{figure}

\begin{figure}[!t]
\centerline{\includegraphics[width=10.5 cm]{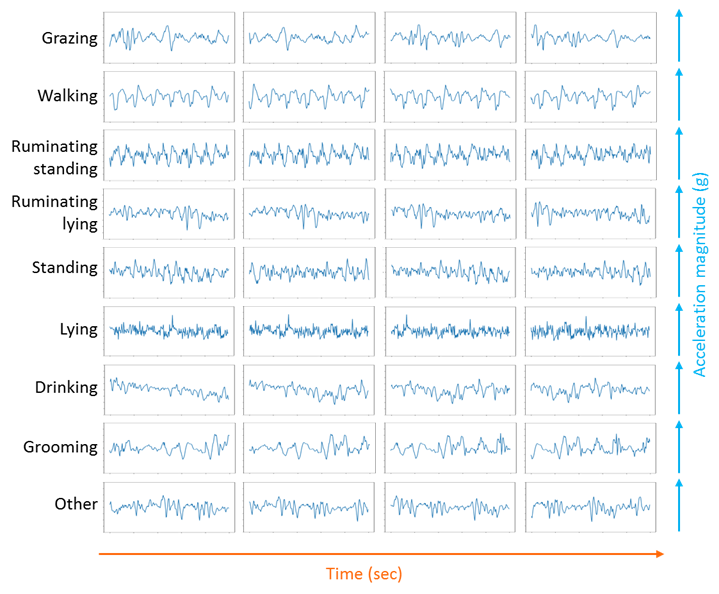}}
\vspace*{4pt}
\caption{Time-series representation of eight consecutive seconds of accelerations signal (z axis) for different activity classes. 
}
\label{fig:TSdiffLabel}
\end{figure}

Figure~\ref{fig:TFDdiffLabel} shows the spectrograms of four consecutive 5 second windows (with 80\% overlap) of the acceleration data. This is shown for the nine different cattle activities. 
For example, in the top row, we see the spectrograms of four windows of Grazing activity. The x axis of each spectrogram represents time, and the y axis represents frequency. The colour indicates the energy density of the acceleration signal, ranging from red (highest) to dark blue (lowest).
What is interesting to see is that each of the eight activities exhibits a unique spectrogram pattern, which is reasonably consistent across the four consecutive time windows, and quite distinct from the other activities. 
In contrast, Figure~\ref{fig:TSdiffLabel} shows exactly the same data, but in time-series representation. Visual inspection of the data in this representation shows much less distinguishing features among the activities. 
For example, while ``Standing" and ``Ruminating Lying" look relatively similar in their respective time-series representation, they exhibit a very different and more clearly distinguishable pattern in the spectrogram representation.

Intuitively, this indicates the potential that a DNN classifier will also be able to more easily distinguish between the different activities, if presented with the spectrogram representation of the sensor data. 
We will test this hypothesis in the remainder of this paper.

%
\subsection{TFD Classifier Architecture}
\begin{figure}[!ht]
\centerline{\includegraphics[width=\textwidth]{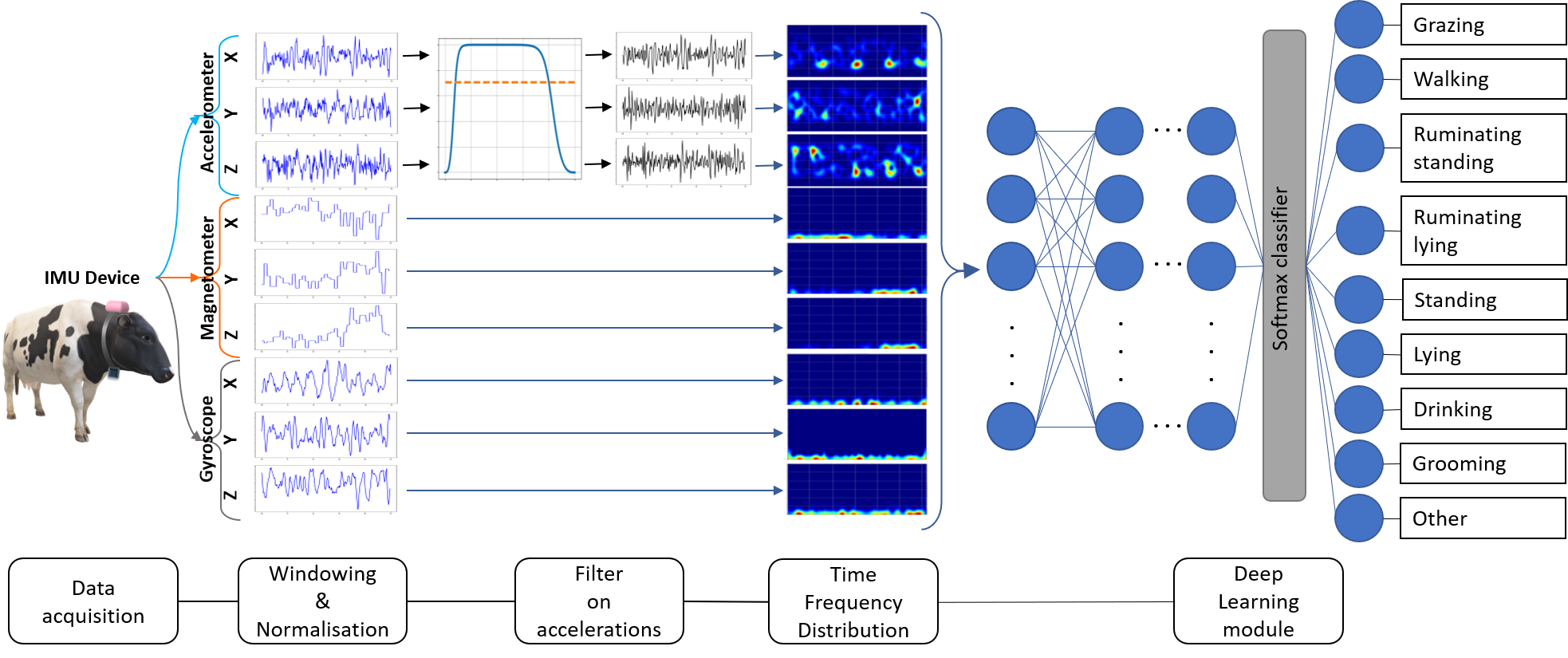}}
\vspace*{8pt}
\caption{Schematic illustration of time-frequency-based activity classification.}
\label{fig:TFDarchitecture}
\end{figure}

The updated procedure and classification architecture with time-frequency representation of the input signals is shown in Figure~\ref{fig:TFDarchitecture}. 
The procedure is similar to that of time-domain-based classification (explained in Section~\ref{ActivityClassificationArchitecture}), except that the data is represented in the joint time-frequency domain. 
The transformation from the time-series representation to time-frequency representation is performed after the windowing, normalisation and filtering, just before the data is fed into the DNN classifier.

\subsection{Results for TFD-based Classification}
The proposed TFD-based classification approach is validated with the Stratified 10-fold cross-validation (SCV) method, as explained in Section~\ref{validation time}.
%
To evaluate the impact of TFD-based data representation on classification performance, we used the same time window sizes as used for time-domain-based classification (Section~\ref{restimewindow}), i.e. $\Delta T = 5s$ with $\Delta t = 1s$ and $\Delta T = 10s$ with $\Delta t = 2s$.

\begin{figure}[!ht]
\centerline{\includegraphics[width=11cm]{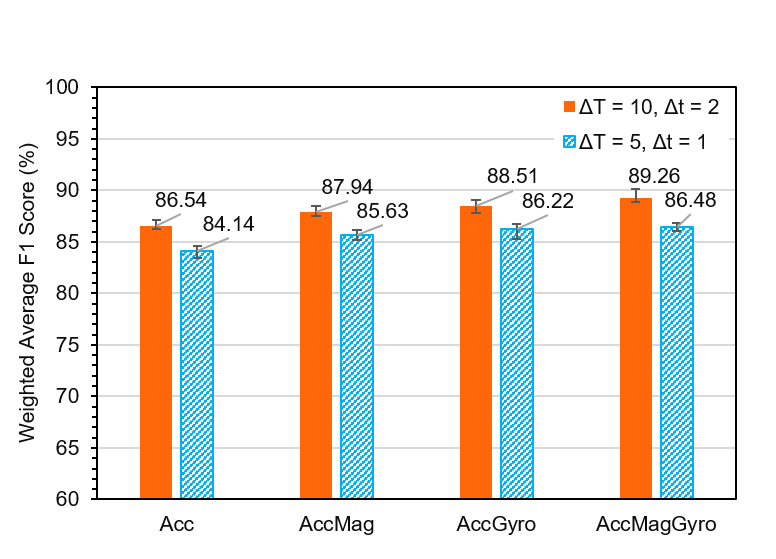}}
\vspace*{8pt}\caption{Classification performance for TFD data representation for different sets of sensor modalities.}
\label{fig:TFDFusionSensors}
\end{figure}

Initially, the fusion of different sensor modalities is explored for the TFD representation, and the results are shown in Figure~\ref{fig:TFDFusionSensors}. 
The best performance of $89.26\%$ is achieved for the case of $\Delta T = 10s$ and $\Delta t = 2s$, with the fusion of all sensor modalities. This is significantly higher than the maximum of $73.55\%$ achieved using time-series data representation.
Furthermore, we observe that adding additional data from different sensor modalities improves the $F_1$ Score by up to $3\%$, in clear contrast to the time-series case.




\begin{figure}[!ht]
\centerline{\includegraphics[width=\textwidth/2]{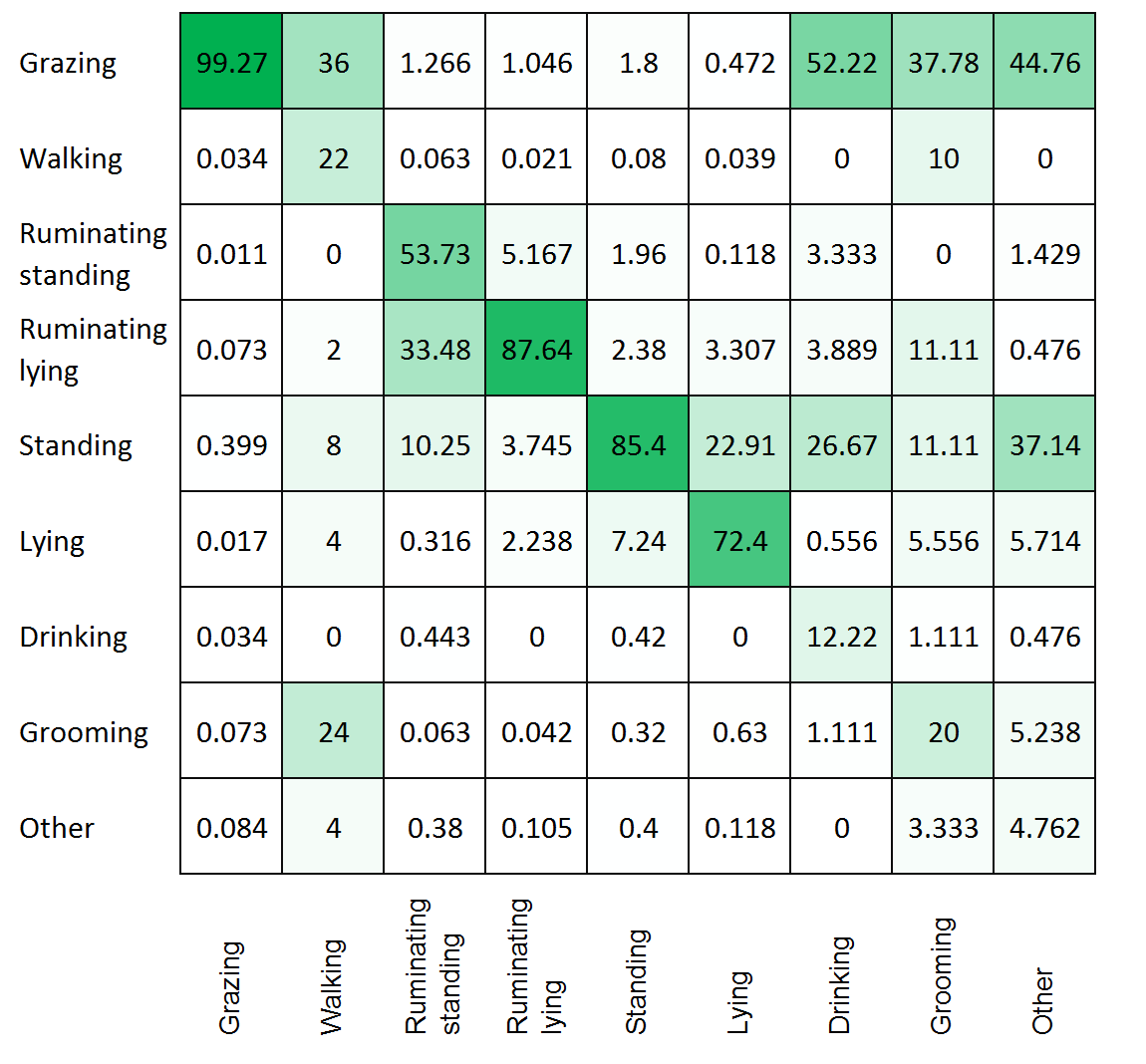}}
\vspace*{8pt}
\caption{Confusion matrix for classification based on TFD data representation, for $\Delta T = 10s$ and $\Delta t = 2s$, based on the combination of accelerometer, magnetometer and gyroscope sensor data, using the SCV method.}
\label{fig:TFDConfusionMat}
\end{figure}

Figure~\ref{fig:TFDConfusionMat} shows the confusion matrix for the TFD-based classification for $\Delta T = 10s$ and $\Delta t = 2s$, and the combination of accelerometer, magnetometer and gyroscope data. 
%
%
%
%
While there are still a significant number of mis-classifications, we observe a clear improvement in classification accuracy for all activity classes compared to the time-series results shown in Figure~\ref{fig:TimeConfusionMat}.


\section{Summary Comparison of Time-Series and TFD-based Classification }\label{Comptfdtim}
In this section, we provide a summary comparison of the two considered data representation approaches, when used in conjunction with the DNN classifier specified in Section~\ref{DNN}. In Section~\ref{TimevsTFD_SCV}, we provide the comparison based on the SCV (10-fold cross validation) method, which we used so far in this paper, and which is used most commonly in the literature in this context. In addition, in Section~\ref{TimevsTFD_LOCV} we also present an evaluation and comparison of both approaches using the leave-one-out cross-validation (LOOCV) method.

\subsection{Comparison based on SCV Method}\label{TimevsTFD_SCV}
\begin{figure}[!ht]
\centerline{\includegraphics[width=14cm]{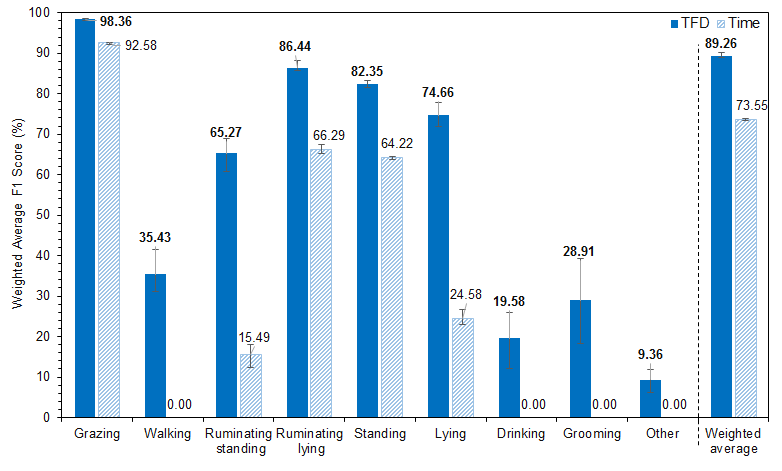}}
\vspace*{8pt}
\caption{Comparison of classification performance for time-series versus TFD sensor data representation using the SCV method, with $\Delta T = 10s$ and $\Delta t = 2s$.}
\label{fig:SCV_TSvsTFD}
\end{figure}

Figure~\ref{fig:SCV_TSvsTFD} compares the $F_1$ Score for the individual cattle activity classes, for both the TFD and time-domain sensor data representation. On the far right, the weighted (average) $F_1$ Scores for both methods are also shown. We show the best combination of sensor modalities for each case, i.e. accelerometer, magnetometer and gyroscope data for TFD representation, and acceleration data only for time-domain representation.  In both cases, the same window size and stride were used, i.e. $\Delta T = 10s$ with $\Delta t = 2s$. 

Consistent with the results shown in the confusion matrices (Figure~\ref{fig:TimeConfusionMat} and ~\ref{fig:TFDConfusionMat}), Figure~\ref{fig:SCV_TSvsTFD} shows that the TFD-based data representation achieves a consistently higher $F_1$ Score across the board for all activity classes. While for some classes, such as ``Walking", ``Drinking" and ``Grooming", the $F_1$ Score is still relatively low, it is significantly higher than the corresponding values achieved by the time-domain approach.
In terms of the weighted (average) $F_1$ Score, the TFD-based approach achieves a more than $15\%$ higher value than the time-domain representation approach, i.e. $89.26\%$ versus $73.55\%$

\subsection{Comparison based on LOOCV Method}\label{TimevsTFD_LOCV}
In addition to a comparison based on the SCV method, we also consider the Leave One Out Cross Validation (LOOCV) method. While computationally expensive, the LOOCV method provides a reliable and unbiased estimate of model performance~\cite{Alpaydin2014}.
In our case, we implement LOOCV by leaving out all the data from one particular animal when training the model, and the data from the unseen animal is used as the test set. We train the model 10 times, each time with leaving out the data of a different animal (of the total population of 10). The performance metrics are calculated as the weighted average over the 10 iterations.



\begin{figure}[!ht]
\centerline{\includegraphics[width=13cm]{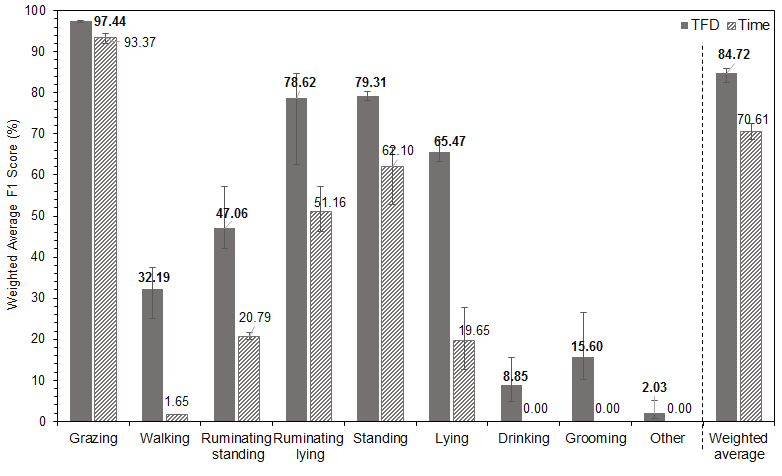}}
\vspace*{8pt}
\caption{Comparison of the classification performance for time-series versus TFD sensor data representation using the LOACV method, with $\Delta T = 5s$ and $\Delta t = 1s$.}
\label{fig:LOV_TSvsTFD}
\end{figure}

Figure~\ref{fig:LOV_TSvsTFD} shows the comparison of the classification performance  based on the LOOCV method (with $\Delta T = 5s$ and $\Delta t = 1s$) \footnote{We chose these parameters for the case of LOOCV since they resulted in a slightly higher classification performance than $\Delta T = 10s$ and $\Delta t = 2s$, which achieved the best performance for the case of SCV.} for fusion of all sensor modalities in joint time-frequency data representation, and accelerations only for time-domain representation.
We observe that the $F_1$ Scores are generally lower with LOOCV for both methods. However, in general, we see a similar relative improvement of TDF-based over time-domain-based data representation. In terms of the average $F_1$ Score, LOOCV shows 
$84.72\%$ vs $70.61\%$, which is consistent with the SCV results of $89.26\%$ vs $73.55\%$.
In summary, the somewhat more stringent evaluation of the LOOCV approach confirms our earlier findings that the TFD-based data representations achieves a significantly higher classification performance than a time-domain representation.



\section{Comparison with State-of-The-Art Algorithms}\label{comparison}
\begin{figure}[!ht]
\centerline{\includegraphics[width=15.5cm]{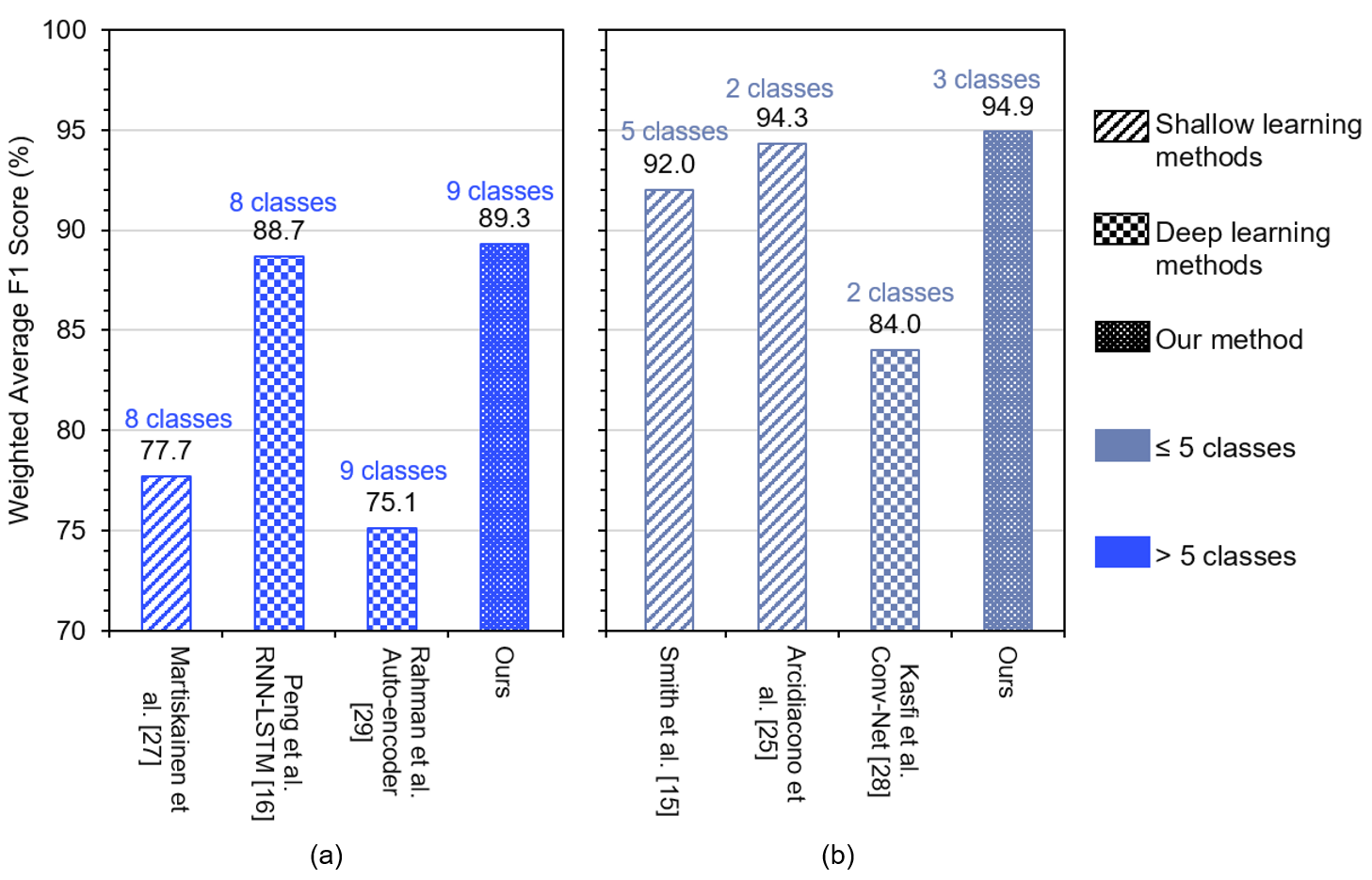}}
\vspace*{5pt}
\caption{Comparison of our proposed approach with state-of-the-art cattle activity classifiers, for $>$ 5 activity classes (a), and $\leq$ 5 activity classes (b).} 
\label{fig:comparelit}
\end{figure}

This section provides a comparison of the cattle activity classification performance of our proposed approach with the best reported results in the literature, to the best of our knowledge. 
Since the classification performance depends significantly on the number of considered activity classes, we group the reported results into two categories, i.e. works with more than 5 classes, and 5 or fewer classes.
Figure~\ref{fig:comparelit}(a)  shows the reported $F_1$ Scores of the 3 best performing cattle activity classifiers for a high number of activity classes, compared to our proposed method.  The shading of the bars indicates the type of classifier, i.e. Shallow vs Deep Learning. For a fair comparison, we used the results based on the SCV method, since this is the method used in the considered related works. 
Peng et al.~\cite{RN131} (second bar from left) reported the highest $F_1$ Score to date of $88.7\%$, using an LSTM-RNN classifier. 
Our own classifier (rightmost bar), achieves the score of $89.3\%$, for 9 activity classes compared to 8 classes of~\cite{RN131}.

Figure~\ref{fig:comparelit}(b), shows the top 3 classification performance results for 5 or fewer activity classes, compared to our approach. For this comparison, we performed an experiment where we only selected data samples from the 3 most populated activity classes of ``Grazing", ``combination of Ruminating lying and Ruminating standing", and ``Standing", and used this reduced dataset for training and evaluation of our proposed activity classifier, as done in~\cite{RN262}. The figure shows that our approach achieves the highest $F_1$ Score of $94.9\%$ for 3 classes, compared to $94.3\%$ of Arcidiacono et al.~\cite{RN147} with only 2 classes. 
In summary, our proposed approach outperforms the best performing cattle activity classifiers reported in the literature, both for a small and large number of activity classes.

The key focus of this paper is the exploration of the TFD-based data representation.
It is therefore important to note that the  improvement over the state-of-the-art has been achieved with a only very basic Neural Network architecture, and with no systematic exploration and optimisation of hyper parameters, in contrast to related works such as~\cite{RN131}.
This demonstrates the predictive power of the TFD-based data representation in this application context, and points to further potential performance improvements, if combined with an optimal choice of Neural Network architecture and hyper parameters. Such explorations represent promising scope for future work.

\section{Varying Spectrogram Resolution}\label{compcost}

In the previous section, we have shown that TFD-based data representation, combined with a relatively simple DNN classifier, can outperform state of the art cattle activity classifiers. 
In this section, we explore another aspect of this approach, i.e. the impact of the spectrogram resolution on the model size, computational cost (inference time), as well as the classification performance.

As shown in  Figure~\ref{fig:netmodel}, the size of the TFD representation of the sensor data, and hence the resolution of the spectrogram,  determines the size of the input layer of our DNN, and to a large extent the overall model size. 
%
 %
%
 \begin{figure}[!ht]
\centerline{\includegraphics[width=7cm]{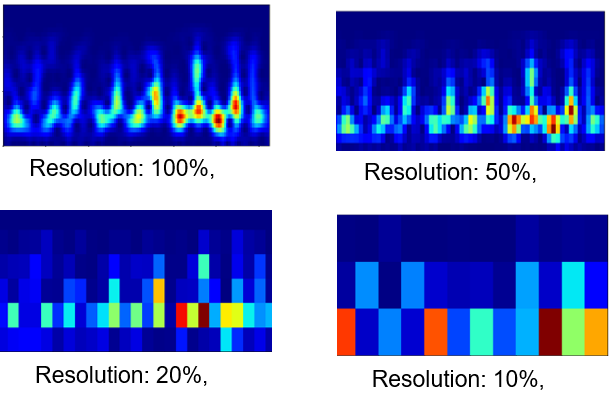}}
\vspace*{8pt}
\caption{Spectrogram data representation with different resolutions, 100\%, 50\%, 20\% and 10\%.}
\label{fig:TFD_ResDown}
\end{figure}
As an illustration, Figure~\ref{fig:TFD_ResDown} shows a sample spectrogram computed from our cattle activity dataset, with four different resolutions, i.e. 100\%, 50\%, 20\% and 10\%. 
A resolution of 100\% corresponds to the full resolution which was used for the classification experiments discussed so far in this paper.
A resolution of 50\% has half the resolution in each of the two dimensions, and hence has a quarter of the `pixels' of the full resolution. Correspondingly, the 10\% resolution has 1\% of the size of the full resolution. The reduced resolutions were computed via downsampling followed by ``bi-cubic interpolation"~\cite{bicubic}.

For each of the four spectrogram resolutions, Table   \ref{tab:compcost} shows the measured inference time (of a single activity classification), the model size in terms of the number of model parameters, as well as the corresponding average $F_1$ Score \footnote{The inference time was measured on an Intel$^{\tiny{\textregistered}}$ Core$^{\tiny{TM}}$ i7-8700 CPU, with 3.2 GHz and 64GB RAM.}. 
%
\begin{table}[t!]
\centering
\caption{Inference time, model size and classification performance for different spectrogram resolutions.}
\small
    \begin{tabular}{cccc}
    \toprule
    \multicolumn{1}{m{7.355em}}{Resolution(\%)} & \multicolumn{1}{m{7.355em}}{Inference\newline{} Time ($\mu$s)} & \multicolumn{1}{m{10.355em}}{Model Parameters\newline{} $\times 10^6$} & \multicolumn{1}{m{10.355em}}{Performance\newline{}(\%Average $F_1$ Score)} \\
    \midrule
    100   & 406.19 & 8.94  & 89.3 \\
    50    & 143.73 & 2.24  & 89.1 \\
    20    & 50.34 & 0.36  & 88.8 \\
    10    & 42.33 & 0.10   & 87.4 \\
    \bottomrule
    \end{tabular}%
  \label{tab:compcost}%

\end{table}%
As expected, the number of model parameters roughly corresponds to the reduction of the squared (one-dimensional) resolution.
As also expected, the inference time reduces with the reduction in model size.
What is surprising though is the fact that the classification performance only very minimally decreases with the reduction in spectrogram resolution. 
For example, for a reduction of the resolution from 100\% to 50\%, and hence a reduction of the model size by a factor of 4, the $F_1$ Score is only reduced by 0.02\%. Even with a quite drastic reduction of the resolution to 10\%, and hence a reduction of the model size to a mere 1\% of its original size, we are sacrificing less than 1\% in terms of the $F_1$ Score. 

This demonstrates the ability of the TFD data representation to significantly `compress' the sensor data without loosing much of its predictive power in the context of cattle activity classification.
This makes TFD a very promising approach for scenarios where deep learning-based activity classification needs to be done at the edge, on embedded devices with limited compute power, energy and communications capacity~\cite{RN801,RN808}.
The exploration of TFD-based data representation for classification tasks at the edge, i.e. on highly resource-constrained IoT devices represents promising direction for future research. 

\section{Conclusions}\label{Conclusion}


In this paper we presented a new approach for cattle activity classification which explores a joint time-frequency domain (TFD) sensor data representation, and combines it with a basic feed-forward Deep Neural Network classifier with four hidden layers. 
Our experimental evaluations are based on a realistic dataset consisting of over 3 million sensor readings from a tri-axial IMU sensor  (accelerometer, magnetometer and gyroscope), attached to collar tags of ten individual dairy cows, collected and manually labelled over a period of one month. 

We compared the classification performance of our DNN classifier, using both TFD and time-domain data representation, and we have shown that the TFD approach significantly outperforms the time-domain representation. 
We further showed that our new TFD-based activity classifier outperforms the best cattle activity classification results reported in the literature, to the best of our knowledge. This includes both shallow and deep learning methods, for both a small and large number of activity classes. 
As mentioned above, the fact that these results were achieved with a very basic DNN classifier demonstrates the power and potential of the TFD data representation for the considered application. It can be expected that with a more advanced neural network architecture and a more optimal choice of hyper-parameters, even better results can be achieved. This represents scope for promising future research.  

Finally, we presented the results of our preliminary exploration of the scalability properties of the TFD data representation. Our results show that the TFD (or spectrogram) representation can be drastically scaled down and compressed, resulting in a significant reduction of the model size and computational cost, with only a very minimal reduction in classification performance. This trade-off of classification accuracy and computational (and energy) cost represents another promising avenue for future work, in particular in an IoT context.

\section*{}
\bibliographystyle{IEEEtran} 
\bibliography{refs.bib} 

\begin{thebibliography}{10}
\providecommand{\url}[1]{#1}
\csname url@samestyle\endcsname
\providecommand{\newblock}{\relax}
\providecommand{\bibinfo}[2]{#2}
\providecommand{\BIBentrySTDinterwordspacing}{\spaceskip=0pt\relax}
\providecommand{\BIBentryALTinterwordstretchfactor}{4}
\providecommand{\BIBentryALTinterwordspacing}{\spaceskip=\fontdimen2\font plus
\BIBentryALTinterwordstretchfactor\fontdimen3\font minus
  \fontdimen4\font\relax}
\providecommand{\BIBforeignlanguage}[2]{{%
\expandafter\ifx\csname l@#1\endcsname\relax
\typeout{** WARNING: IEEEtran.bst: No hyphenation pattern has been}%
\typeout{** loaded for the language `#1'. Using the pattern for}%
\typeout{** the default language instead.}%
\else
\language=\csname l@#1\endcsname
\fi
#2}}
\providecommand{\BIBdecl}{\relax}
\BIBdecl

\bibitem{RN256}
OECD/FAO, \emph{OECD/FAO (2018)}.\hskip 1em plus 0.5em minus 0.4em\relax OECD
  Publishing, Paris/Food and Agriculture Organization of the United Nations,
  Rome, 2018, book section 6-7, pp. 149--174.

\bibitem{RN247}
\BIBentryALTinterwordspacing
``{GLOBAL SNAPSHOT {$|$} BEEF},'' {Meat \& Livestock, Australia (mla)}, Report,
  January 2019. [Online]. Available:
  \url{https://www.mla.com.au/globalassets/mla-corporate/prices--markets/documents/os-markets/export-statistics/jan-2019-snapshots/global-beef-snapshot-jan2019.pdf}
\BIBentrySTDinterwordspacing

\bibitem{RN253}
\BIBentryALTinterwordspacing
A.~Nasirahmadi, S.~A. Edwards, and B.~Sturm, ``Implementation of machine vision
  for detecting behaviour of cattle and pigs,'' \emph{Livestock Science}, vol.
  202, no. May, pp. 25--38, 2017. [Online]. Available:
  \url{http://dx.doi.org/10.1016/j.livsci.2017.05.014}
\BIBentrySTDinterwordspacing

\bibitem{RN618}
J.~Moran and R.~Doyle, \emph{Cow Talk}.\hskip 1em plus 0.5em minus 0.4em\relax
  CSIRO Publishing, 2015.

\bibitem{RN316}
M.~R. Borchers, Y.~M. Chang, K.~L. Proudfoot, B.~A. Wadsworth, A.~E. Stone, and
  J.~M. Bewley, ``Machine-learning-based calving prediction from activity,
  lying, and ruminating behaviors in dairy cattle,'' \emph{Journal of Dairy
  Science}, vol. 100, no.~7, pp. 5664--5674, 2017.

\bibitem{RN484}
C.~Phillips, \emph{Cattle Behaviour and Welfare}.\hskip 1em plus 0.5em minus
  0.4em\relax Malden, MA, USA: Blackwell Science Ltd, 2002, book section~2, pp.
  10--21.

\bibitem{RN814}
\BIBentryALTinterwordspacing
{Leibniz Institute of Agricultural Development in Transition Economies(IAMO)},
  ``Top 10 australia's biggest cattle stations,'' 2018. [Online]. Available:
  \url{https://www.largescaleagriculture.com/home/news-details/top-10-australias-biggest-cattle-stations/}
\BIBentrySTDinterwordspacing

\bibitem{RN262}
A.~Rahman, D.~V. Smith, B.~Little, A.~B. Ingham, P.~L. Greenwood, and G.~J.
  Bishop-Hurley, ``Cattle behaviour classification from collar, halter, and ear
  tag sensors,'' \emph{Information Processing in Agriculture}, vol.~5, no.~1,
  pp. 124--133, 2018.

\bibitem{RN272}
M.~S. Shahriar, D.~Smith, A.~Rahman, M.~Freeman, J.~Hills, R.~Rawnsley,
  D.~Henry, and G.~Bishop-Hurley, ``Detecting heat events in dairy cows using
  accelerometers and unsupervised learning,'' \emph{Computers and Electronics
  in Agriculture}, vol. 128, pp. 20--26, 2016.

\bibitem{RN106}
R.~Dutta, D.~Smith, R.~Rawnsley, G.~Bishop-Hurley, and J.~Hills, ``Cattle
  behaviour classification using 3-axis collar sensor and multi-classifier
  pattern recognition,'' \emph{Proceedings of IEEE Sensors}, vol. 2014-Decem,
  no. December, pp. 1272--1275, 2014.

\bibitem{RN151}
L.~A. González, K.~S. Schwartzkopf-Genswein, N.~A. Caulkett, E.~Janzen, T.~A.
  McAllister, E.~Fierheller, A.~L. Schaefer, D.~B. Haley, J.~M. Stookey, and
  S.~Hendrick, ``Pain mitigation after band castration of beef calves and its
  effects on performance, behavior, escherichia coli, and salivary cortisol1,''
  \emph{Journal of Animal Science}, vol.~88, no.~2, pp. 802--810, 2010.

\bibitem{RN110}
R.~N. Handcock, D.~L. Swain, G.~J. Bishop-Hurley, K.~P. Patison, T.~Wark,
  P.~Valencia, P.~Corke, and C.~J. O'Neill, ``Monitoring animal behaviour and
  environmental interactions using wireless sensor networks, gps collars and
  satellite remote sensing,'' \emph{Sensors}, vol.~9, no.~5, pp. 3586--3603,
  2009.

\bibitem{RN155}
D.~M. Weary, J.~M. Huzzey, and M.~A.~G. von Keyserlingk, ``Board-invited
  review: Using behavior to predict and identify ill health in animals,''
  \emph{Journal of Animal Science}, vol.~87, no.~2, pp. 770--777, 2009.

\bibitem{RN103}
R.~Dutta, D.~Smith, R.~Rawnsley, G.~Bishop-Hurley, J.~Hills, G.~Timms, and
  D.~Henry, ``Dynamic cattle behavioural classification using supervised
  ensemble classifiers,'' \emph{Computers and Electronics in Agriculture}, vol.
  111, pp. 18--28, 2015.

\bibitem{RN4}
D.~Smith, A.~Rahman, G.~J. Bishop-Hurley, J.~Hills, S.~Shahriar, D.~Henry, and
  R.~Rawnsley, ``Behavior classification of cows fitted with motion collars:
  Decomposing multi-class classification into a set of binary problems,''
  \emph{Computers and Electronics in Agriculture}, vol. 131, pp. 40--50, 2016.

\bibitem{RN131}
Y.~Peng, N.~Kondo, T.~Fujiura, T.~Suzuki, Wulandari, H.~Yoshioka, and
  E.~Itoyama, ``Classification of multiple cattle behavior patterns using a
  recurrent neural network with long short-term memory and inertial measurement
  units,'' \emph{Computers and Electronics in Agriculture}, vol. 157, pp.
  247--253, 2019.

\bibitem{RN144}
\BIBentryALTinterwordspacing
J.~Wang, Y.~Chen, S.~Hao, X.~Peng, and L.~Hu, ``Deep learning for sensor-based
  activity recognition: A survey,'' \emph{Pattern Recognition Letters}, vol.
  119, pp. 3--11, 2019. [Online]. Available:
  \url{\url{https://doi.org/10.1016/j.patrec.2018.02.010}}
\BIBentrySTDinterwordspacing

\bibitem{RN157}
B.~D. Rouhani, A.~Mirhoseini, and F.~Koushanfar, ``Tinydl: Just-in-time deep
  learning solution for constrained embedded systems,'' \emph{IEEE
  International Symposium on Circuits and Systems (Iscas)}, pp. 452--455, 2017.

\bibitem{RN133}
A.~{Rahman}, D.~{Smith}, J.~{Hills}, G.~{Bishop-Hurley}, D.~{Henry}, and
  R.~{Rawnsley}, ``A comparison of autoencoder and statistical features for
  cattle behaviour classification,'' in \emph{International Joint Conference on
  Neural Networks (IJCNN)}, July 2016, pp. 2954--2960.

\bibitem{RN311}
D.~P. Kumar, T.~Amgoth, C.~Sekhara, and R.~Annavarapu, ``Machine learning
  algorithms for wireless sensor networks : A survey,'' \emph{Information
  Fusion}, vol.~49, no. April 2018, pp. 1--25, 2019.

\bibitem{RN479}
Y.~Kim and T.~Moon, ``Human detection and activity classification based on
  micro-doppler signatures using deep convolutional neural networks,''
  \emph{IEEE Geoscience and Remote Sensing Letters}, vol.~13, no.~1, pp. 8--12,
  2016.

\bibitem{RN483}
D.~Ravi, C.~Wong, B.~Lo, and G.~Yang, ``Deep learning for human activity
  recognition: A resource efficient implementation on low-power devices,'' in
  \emph{2016 IEEE 13th International Conference on Wearable and Implantable
  Body Sensor Networks (BSN)}, 2016, Conference Proceedings, pp. 71--76.

\bibitem{fetmov}
S.~{Layeghy}, G.~{Azemi}, P.~{Colditz}, and B.~{Boashash}, ``Classification of
  fetal movement accelerometry through time-frequency features,'' in \emph{8th
  International Conference on Signal Processing and Communication Systems
  (ICSPCS)}, Dec 2014, pp. 1--6.

\bibitem{RN265}
D.~Ravi, C.~Wong, B.~Lo, and G.-Z. Yang, ``A deep learning approach to on-node
  sensor data analytics for mobile or wearable devices,'' \emph{IEEE Journal of
  Biomedical and Health Informatics}, vol.~21, no.~1, pp. 56--64, 2017.

\bibitem{RN162}
M.~A. Alsheikh, A.~Selim, D.~Niyato, L.~Doyle, S.~Lin, and H.-P. Tan, ``Deep
  activity recognition models with triaxial accelerometers,'' in
  \emph{Workshops at the Thirtieth AAAI Conference on Artificial Intelligence},
  2016.

\bibitem{networkref}
Z.~{Wang}, W.~{Yan}, and T.~{Oates}, ``Time series classification from scratch
  with deep neural networks: A strong baseline,'' in \emph{2017 International
  Joint Conference on Neural Networks (IJCNN)}, 2017, pp. 1578--1585.

\bibitem{RN978}
\BIBentryALTinterwordspacing
``A rough-granular approach to the imbalanced data classification problem,''
  \emph{Applied Soft Computing}, vol.~83, p. 105607, 2019. [Online]. Available:
  \url{http://www.sciencedirect.com/science/article/pii/S1568494619303874}
\BIBentrySTDinterwordspacing

\bibitem{RN801}
\BIBentryALTinterwordspacing
C.~TAG, ``What is ceres tag,'' 2019. [Online]. Available:
  \url{\url{https://www.cerestag.com/}}
\BIBentrySTDinterwordspacing

\bibitem{RN808}
\BIBentryALTinterwordspacing
CSIRO, ``Ceres tag: smart ear tags for livestock,'' 2019. [Online]. Available:
  \url{\url{https://www.csiro.au/en/Research/AF/Areas/Livestock/Ceres-Tag}}
\BIBentrySTDinterwordspacing

\bibitem{RN313}
A.~L.~H. Andriamandroso, F.~Lebeau, Y.~Beckers, E.~Froidmont, I.~Dufrasne,
  B.~Heinesch, P.~Dumortier, G.~Blanchy, Y.~Blaise, and J.~Bindelle,
  ``Development of an open-source algorithm based on inertial measurement units
  (imu) of a smartphone to detect cattle grass intake and ruminating
  behaviors,'' \emph{Computers and Electronics in Agriculture}, vol. 139, pp.
  126--137, 2017.

\bibitem{RN315}
J.~A. Vázquez~Diosdado, Z.~E. Barker, H.~R. Hodges, J.~R. Amory, D.~P. Croft,
  N.~J. Bell, and E.~A. Codling, ``Classification of behaviour in housed dairy
  cows using an accelerometer-based activity monitoring system,'' \emph{Animal
  Biotelemetry}, vol.~3, no.~1, p.~15, 2015.

\bibitem{RN147}
C.~Arcidiacono, S.~M.~C. Porto, M.~Mancino, and G.~Cascone, ``Development of a
  threshold-based classifier for real-time recognition of cow feeding and
  standing behavioural activities from accelerometer data,'' \emph{Computers
  and Electronics in Agriculture}, vol. 134, pp. 124--134, 2017.

\bibitem{RN218}
V.~Sturm, D.~Efrosinin, N.~Efrosinina, L.~Roland, M.~Iwersen, M.~Drillich, and
  W.~Auer, ``A chaos theoretic approach to animal activity recognition,''
  \emph{Journal of Mathematical Sciences}, vol. 237, no.~5, pp. 730--743, 2019.

\bibitem{RN821}
P.~Martiskainen, M.~Jarvinen, J.~P. Skon, J.~Tiirikainen, M.~Kolehmainen, and
  J.~Mononen, ``Cow behaviour pattern recognition using a three-dimensional
  accelerometer and support vector machines,'' \emph{Applied Animal Behaviour
  Science}, vol. 119, no. 1-2, pp. 32--38, 2009.

\bibitem{RN221}
K.~T. Kasfi, A.~Hellicar, and A.~Rahman, ``Convolutional neural network for
  time series cattle behaviour classification,'' in \emph{Time Series Analytics
  and Applications (TSAA)}.\hskip 1em plus 0.5em minus 0.4em\relax ACM Press,
  2016, Conference Proceedings, pp. 8--12.

\bibitem{RN263}
A.~Rahman, D.~Smith, J.~Hills, G.~Bishop-Hurley, D.~Henry, and R.~Rawnsley, ``A
  comparison of autoencoder and statistical features for cattle behaviour
  classification,'' in \emph{2016 International Joint Conference on Neural
  Networks (IJCNN)}, 2016, Conference Proceedings, pp. 2954--2960.

\bibitem{RN477}
H.~F. Nweke, Y.~W. Teh, G.~Mujtaba, and M.~A. Al-garadi, ``Data fusion and
  multiple classifier systems for human activity detection and health
  monitoring: Review and open research directions,'' \emph{Information Fusion},
  vol.~46, pp. 147--170, 2019.

\bibitem{Chang2015}
\BIBentryALTinterwordspacing
A.~X.~M. Chang, B.~Martini, and E.~Culurciello, ``{Recurrent Neural Networks
  Hardware Implementation on FPGA},'' \emph{ArXiv}, Nov 2015. [Online].
  Available: \url{http://arxiv.org/abs/1511.05552}
\BIBentrySTDinterwordspacing

\bibitem{myPhdForum}
S.~{Hosseininoorbin}, ``Phd forum abstract: Activity classification at the
  edge,'' in \emph{2020 19th ACM/IEEE International Conference on Information
  Processing in Sensor Networks (IPSN)}, 2020, pp. 369--370.

\bibitem{RN826}
H.~J. Luinge and P.~H. Veltink, ``Inclination measurement of human movement
  using a 3d accelerometer with autocalibration,'' \emph{IEEE Transactions on
  Neural Systems and Rehabilitation Engineering}, vol.~12, no.~1, pp. 112--121,
  2004.

\bibitem{RN464}
F.~Ertam and G.~Aydin, ``Data classification with deep learning using
  tensorflow,'' in \emph{International Conference on Computer Science and
  Engineering (Ubmk)}, 2017, Conference Proceedings, pp. 755--758.

\bibitem{RN465}
V.~Nair and G.~E. Hinton, ``Rectified linear units improve restricted boltzmann
  machines,'' in \emph{Proceedings of the 27th international conference on
  machine learning (ICML-10)}, 2010, Conference Proceedings, pp. 807--814.

\bibitem{RN466}
H.~Ismail~Fawaz, G.~Forestier, J.~Weber, L.~Idoumghar, and P.-A. Muller, ``Deep
  learning for time series classification: a review,'' \emph{Data Mining and
  Knowledge Discovery}, vol.~33, no.~4, pp. 917--963, 2019.

\bibitem{Murphy2012}
K.~P. Murphy, ``{Generalized linear models and the exponential family},'' in
  \emph{Machine learning: a probabilistic perspective (adaptive computation and
  machine learning series)}.\hskip 1em plus 0.5em minus 0.4em\relax MIT Press,
  2012, ch.~9, pp. 301--302.

\bibitem{keras}
F.~Chollet \emph{et~al.}, ``Keras,'' \url{https://keras.io}, 2015.

\bibitem{tensorflow2015}
\BIBentryALTinterwordspacing
M.~Abadi \emph{et~al.}, ``{TensorFlow}: Large-scale machine learning on
  heterogeneous systems,'' 2015, software available from tensorflow.org.
  [Online]. Available: \url{https://www.tensorflow.org/}
\BIBentrySTDinterwordspacing

\bibitem{kingma2014}
\BIBentryALTinterwordspacing
D.~P. Kingma and J.~Ba, ``Adam: A method for stochastic optimization,''
  \emph{ArXiv}, Dec 2015. [Online]. Available:
  \url{http://arxiv.org/abs/1412.6980}
\BIBentrySTDinterwordspacing

\bibitem{RN827}
N.~Ketkar, \emph{Deep Learning with Python}.\hskip 1em plus 0.5em minus
  0.4em\relax Apress, 2017, book section~8, p. 125.

\bibitem{Alpaydin2014}
E.~Alpaydin, ``{Design and Analysis of Machine Learning Experiments},'' in
  \emph{Introduction to machine learning}.\hskip 1em plus 0.5em minus
  0.4em\relax MIT Press, 2014, ch.~19, pp. 559--560.

\bibitem{bicubic}
P.~Xia, T.~Tahara, T.~Kakue, Y.~Awatsuji, K.~Nishio, S.~Ura, T.~Kubota, and
  O.~Matoba, ``Performance comparison of bilinear interpolation, bicubic
  interpolation, and b-spline interpolation in parallel phase-shifting digital
  holography,'' \emph{Optical review}, vol.~20, no.~2, pp. 193--197, 2013.

\end{thebibliography}

\vspace*{-0.01in}
\noindent
\rule{12.6cm}{.1mm}

\end{document}